\def\doublecolumn{1} 
\if 1\doublecolumn
  \documentclass[journal]{IEEEtran}
\else
  \documentclass[12pt, draftclsnofoot, onecolumn, letterpaper]{IEEEtran}
\fi

\def\blind{1} 

\usepackage[latin9]{inputenc}
\usepackage{algorithm}

\usepackage{algcompatible}

\algblockdefx{FORALLP}{ENDFAP}[1]%
  {\textbf{for }#1 \textbf{do in parallel}}%
  {\textbf{end for}}

\usepackage{array}

\usepackage{bbm}
\usepackage{dsfont}
\usepackage{hyperref}
\usepackage{makecell} 
\usepackage{float}
\usepackage{mathtools}
\usepackage{amsmath}
\usepackage{amsthm}
\usepackage{amssymb}
\usepackage{graphicx}
\usepackage{setspace}
\usepackage{color}
\usepackage{bm}
\usepackage{cases}
\usepackage{tikz}
\usetikzlibrary{calc}
\usepackage{flowchart}
\usepackage{bbm}
\usepackage{adjustbox}
\usepackage{hhline}
\usepackage{makecell}

\usepackage{epsfig}
\usepackage{cite}
\usepackage{graphicx}
\usepackage{lipsum}
\usepackage{multirow}
\usepackage{array}
\usepackage{enumerate}
\usepackage{enumitem}
\usepackage{graphicx}
\usepackage{times}
\usepackage[framemethod=tikz]{mdframed}
\definecolor{cccolor}{rgb}{1,1,1}

\usepackage[caption=false]{subfig}
\newtheorem{theorem}{Theorem}
\newtheorem{lemma}{Lemma}
\newtheorem{assumption}{Assumption}

\newtheorem{proposition}{Proposition}
\newtheorem{definition}{Definition}
\DeclareMathOperator*{\argmin}{argmin}

\usepackage{multicol,multirow,adjustbox}
\usepackage{booktabs}

\newcommand{\eg}[0]{\textit{e.g.}}
\newcommand{\ie}[0]{\textit{i.e.}}

\usepackage[capitalize]{cleveref}
\Crefname{section}{Section}{Sections}
\Crefname{table}{Table}{Tables}
\usepackage{fontawesome}

\makeatletter
\def\@IEEEsectpunct{.\ \,}
\def\paragraph{\@startsection{paragraph}{4}{\z@}{2.0ex}%
{0ex}{\normalfont\normalsize\bfseries}}
\makeatother

\usepackage{pifont,xspace}
\newcommand{\cmark}{\ding{51}\xspace}%
\newcommand{\cmarkg}{\textcolor{lightgray}{\ding{51}}\xspace}%
\newcommand{\xmarkg}{\textcolor{lightgray}{\ding{55}}\xspace}%

\newcounter{problem}
\newcounter{save@equation}
\newcounter{save@problem}

\makeatletter
\newenvironment{problem}
 {\setcounter{problem}{\value{save@problem}}%
  \setcounter{save@equation}{\value{equation}}%
  \let\c@equation\c@problem
  \subequations
  }
 {\endsubequations
  \setcounter{save@problem}{\value{equation}}%
  \setcounter{equation}{\value{save@equation}}%
 }

\begin{document}

\title{
    Fed-ZOE: Communication-Efficient Over-the-Air Federated Learning via Zeroth-Order Estimation
}

\if 1\blind
\author{
Jonggyu~Jang,~\IEEEmembership{Member,~IEEE}, 
Hyeonsu~Lyu,~\IEEEmembership{Student Member,~IEEE},
David~J.~Love,~\IEEEmembership{Fellow,~IEEE},
and 
Hyun~Jong~Yang,~\IEEEmembership{Member,~IEEE}
    \thanks{
    J. Jang is with Institute of New Media and Communications, Seoul National University, Seoul 08826, Republic of Korea, (e-mail: jgjang@snu.ac.kr), and also with the Elmore Family School of Electrical and Computer Engineering, Purdue University, West Lafayette, IN 47907 USA (e-mail: jang255@purdue.edu). 
    H. Lyu is with Department of Electrical Engineering, Pohang University of Science and Technology (POSTECH), Pohang 37673, Republic of Korea, (e-mail: hslyu4@postech.ac.kr).
    D. J. Love is with the Elmore Family School of Electrical and Computer Engineering, Purdue University, West Lafayette, IN 47907 USA (e-mail: djlove@purdue.edu).
    H. J. Yang (corresponding author) is with Department of Electrical and Computer Engineering, Seoul National University, Seoul 08826, Republic of Korea, (e-mail: hjyang@snu.ac.kr).
    }
}
\else
\author{Anonymous Submission}
\fi
\maketitle

\begin{abstract}
    As 6G and beyond networks grow increasingly complex and interconnected, federated learning (FL) emerges as an indispensable paradigm for securely and efficiently leveraging decentralized edge data for AI.
    By virtue of the superposition property of communication signals, over-the-air FL (OtA-FL) achieves constant communication overhead irrespective of the number of edge devices (EDs).
    However, training neural networks over the air still incurs substantial communication costs, as the number of transmitted symbols equals the number of trainable parameters.
    To alleviate this issue, the most straightforward approach is to reduce the number of transmitted symbols by 1) gradient compression and 2) gradient sparsification. 
    Unfortunately, these methods are incompatible with OtA-FL due to the loss of its superposition property.
    In this work, we introduce \textit{federated zeroth-order estimation (Fed-ZOE)}, an efficient framework inspired by the randomized gradient estimator (RGE) commonly used in zeroth-order optimization (ZOO).
    In Fed-ZOE, EDs perform local weight updates as in standard FL, but instead of transmitting full gradient vectors, they send compressed local model update vectors in the form of several scalar-valued inner products between the local model update vectors and random vectors.
    These scalar values enable the parameter server (PS) to reconstruct the gradient using the RGE trick with highly reduced overhead, as well as preserving the superposition property.
    Unlike conventional ZOO leveraging RGE for step-wise gradient descent, Fed-ZOE compresses local model update vectors before transmission, thereby achieving higher accuracy and computational efficiency.
    Numerical evaluations using ResNet-18 on datasets such as CIFAR-10, Tiny-ImageNet, SVHN, CIFAR-100, and Brain-CT demonstrate that Fed-ZOE achieves performance comparable to Fed-OtA while drastically reducing communication costs.
    Notably, the proposed method requires at most 30\% of the communication overhead of traditional approaches, offering a significant advantage for communication-efficient FL.
\end{abstract}

\section{Introduction}
\label{sec:Introduction}

The rapid proliferation of edge devices (EDs)---ranging from IoT sensors to vehicles and drones---has rendered their locally generated data indispensable for training machine learning (ML) models. 
However, directly transmitting such data to a centralized server introduces severe privacy and communication overhead challenges~\cite{ruzomberka2023challenges}, which are set to intensify in 6G environments~\cite{brinton2024key}.
Federated learning (FL) addresses these concerns by enabling EDs to collaboratively train a global model without sharing local data. In FL, EDs train local models using shared weights and transmit only gradient information to a central server, thus ensuring privacy. This paradigm has been successfully adopted in various domains, including medical AI~\cite{dayan2021federated}, recommendation systems~\cite{chatterjee2023federated}, and autonomous driving~\cite{chellapandi2023federated}.

\paragraph*{Over-the-air FL}

Despite the advantages of FL, traditional implementations over wireless networks often rely on orthogonal multiple access (OMA) schemes for communication. These schemes can be inefficient and have limited scalability due to the need for dedicated communication resources for each device~\cite{yang2020federated}.
To address these limitations, the concept of over-the-air computing (OAC) offers a more efficient alternative by enabling nomographic functions---a class of functions that can be decomposed into a sum of component functions $\phi_i$ of local inputs $\mathbf{w}_i$, followed by a global operation $\psi$~\cite{goldenbaum2014nomographic} such as 
\begin{equation}
f(\mathbf{w}_1, \mathbf{w}_2, \dots, \mathbf{w}_N) = \psi \left( \sum_{i=1}^{N} \phi_i(\mathbf{w}_i) \right),
\end{equation}
where $\phi_i$ represents a local transformation at device $i$ and $\psi$ is a global function applied to the sum of transformed inputs.
In the context of \texttt{FedAvg}~\cite{pmlr-v54-mcmahan17a}, the function $f$ becomes a weighted sum of the local updates $\mathbf{w}_i$, aligning perfectly with the model aggregation process in FL where local updates are summed over devices\cite{yang2020federated}.
Specifically, in the simplest case, we have $\phi_i(\mathbf{w}_i) = D_i \mathbf{w}_i$ and $\psi(\mathbf{x}) = \frac{1}{\sum_{i} D_i} \mathbf{x}$, where $D_i$ denotes the number of data samples at the $i$-th edge device.
\begin{table*}[!ht]
    \caption{Previous studies related to communication-efficient federated learning. }
    \centering
    \adjustbox{width=1\linewidth}{
    \begin{tabular}{cccccccc}
    \toprule
        Ref. \# &  MIMO & OtA & Reduced Comm. Overhead & Optimization Order & Large Neural Networks & Convergence Speed & Pub. year \\
    \midrule
        \cite{yang2020federated} & \cmark & \cmark & \xmarkg & First & \xmarkg & \faThumbsOUp\faThumbsOUp & 2020 \\ 
        \cite{jeon2022communication} & \cmark & \xmarkg & \faThumbsOUp~(CS) & First & \xmarkg & \faThumbsOUp & 2022 \\
        \cite{fan20221} & \xmarkg & \cmark & \faThumbsOUp~(CS) & First & \xmarkg & - & 2021 \\ 
        \cite{zhang2022coded} & \xmarkg & \cmark & \faThumbsOUp~(Source Coding) & First & \xmarkg & - & 2023 \\ 
        \cite{oh2024communication} & \xmarkg & \cmark & \faThumbsOUp~(SQ) &  First & \xmarkg & - & 2024 \\ 
        \cite{fang2022communication} & \xmarkg & \xmarkg & - & Zero & \xmarkg & \faThumbsDown & 2022 \\ 
        \cite{malladi2023fine} & \xmarkg & \xmarkg &  \faThumbsOUp\faThumbsOUp~(ZOO) & Zero & \cmark & \faThumbsDown & 2023 \\ 
        \cite{qin2023federated} & \xmarkg & \xmarkg &  \faThumbsOUp\faThumbsOUp~(ZOO) & Zero & \cmark & \faThumbsDown & 2024 \\ 
        \cite{kuo2024federated} & \cmarkg & \cmarkg & \faThumbsOUp~(LoRA) & First & \cmark & \faThumbsOUp~(Fine-tuning-only) & 2024 \\
        \cite{xu2022pruning} & \cmark & \cmark & - (Pruning) & First & \xmarkg & - & 2022 \\ 
        \cite{yang2022over} & \xmarkg & \cmark & - & Second & \xmarkg & - & 2022\\ 
        \midrule
        \textbf{Ours} & \cmark & \cmark & \faThumbsOUp\faThumbsOUp~(RGE) & \textbf{First} & \cmark & \faThumbsOUp\faThumbsOUp & - \\
    \bottomrule
    \multicolumn{8}{l}{* CS: compressed sensing, ZOO: zero-th order optimization, SQ: sparse quantization, RGE: randomized gradient compression. }\\
    \multicolumn{8}{l}{** Convergence speed is relative to the perfect aggregation.}
    \end{tabular}
    }
    \label{tab:existing}
\end{table*}

\paragraph*{Challenges}

Over-the-air federated learning (OtA-FL) has garnered significant research attention due to the advantages offered by OAC\cite{hellstrom2023federated,chellapandi2023federated}. While several studies have explored the feasibility of OtA-FL for large neural networks\cite{chu2022federated}, recent trends reveal a rapid and exponential growth in ML model sizes. For instance, model sizes are reported to increase by approximately 10 times per year, leading to the following critical challenges~\cite{wu2023peta}.

\paragraph*{C1) Communication Overhead}

In OtA-FL, each communication subcarrier corresponds to a single neural network parameter.
Consequently, as the number of parameters grows exponentially with increasing model complexity, the number of symbols exchanged via wireless channels also increases, leading to significant communication overhead.
For instance, consider OtA-FL with a resource block consisting of 12 subcarriers (a sample spacing of $66.7\mu s$\cite{3gpp.38.211}). Then, transmitting $1.1 \times 10^7$ parameters (as in ResNet-18\cite{he2016deep}) would require 61.1 seconds for a single uplink aggregation.
Similarly, during the downlink broadcasting phase with 20 EDs, transmitting $1.1 \times 10^7$ parameters over the communication channel results in a considerable communication overhead, approximately $3.5 \times 10^{8}$ bits. 
Moreover, as proposed in~\cite{hellstrom2023federated}, for enhanced robustness in OtA-FL, fast-fading retransmission of OtA symbols can be employed to mitigate noise through coherence integration.
However, this approach significantly amplifies the communication overhead, potentially extending the duration to hundreds of seconds.

In conventional FL, several approaches have been proposed to reduce communication overhead, including 1) compressed sensing~\cite{jeon2022communication, fan20221}, 2) signSGD~\cite{bernstein2018signsgd,jangrethinking}, 3) gradient sparsification~\cite{oh2024communication, lin2017deep}, and 4) gradient compression~\cite{haddadpour2021federated}. 
The compressed sensing method has been explored in~\cite{jeon2022communication, fan20221}, where it is used for gradient compression.
However, this approach requires iterative computations to converge on the compressed sensing parameters, making it computationally expensive. 
Additionally, its effectiveness has primarily been demonstrated on small neural networks with a limited number of feed-forward layers.
In~\cite{fan20221}, compressed sensing was applied to OtA-FL, but the convergence performance degraded significantly, even with a compression ratio of 20\%.
On the other hand, gradient sparsification/quantization techniques, proposed in~\cite{oh2024communication, lin2017deep, zhang2022coded} often utilize layer-wise sparsification and one-bit quantization combined with error-feedback mechanisms. However, sparsification requires transmitting additional indicators, incurring extra communication costs, and its performance degradation becomes more pronounced as model size increases.

\paragraph*{C2) Accurate Gradient Descent in Zeroth-Order Optimization} 

Zeroth-order optimization (ZOO) methods for FL initially proposed in~\cite{fang2022communication}, but their potential for reducing communication overhead garnered significant research interest in later studies.
In subsequent works~\cite{neto2024communication, malladi2023fine, 8364613}, ZOO was explored as a communication-efficient approach by updating local parameters using directional derivatives in random directions. However, despite its efficiency, ZOO introduces a critical challenge: accurate gradient estimation at the parameter server (PS), essential for achieving reliable convergence.

Additionally, low-rank adaptation (LoRA), a parameter-efficient fine-tuning method, has been explored for communication-efficient FL by transmitting only the low-rank components of model parameters~\cite{kuo2024federated, hu2021lora}.
However, aggregating the low-rank components from multiple EDs breaks the superposition property in OtA-FL, thereby resulting in unintended parameter aggregation.
Furthermore, as LoRA is designed for applying small parameter updates through low-rank adapters, it is unsuitable for training models from scratch, limiting its applicability in certain FL scenarios.

\Cref{tab:existing} summarizes existing studies and provides a comparison with the proposed method.

\paragraph*{Research question}

Motivated by the above challenges of the OtA-FL for large neural networks, we identify the following  research question:
\begin{mdframed}[outerlinecolor=black,outerlinewidth=1pt,linecolor=cccolor,middlelinewidth=1pt,roundcorner=0pt]
  \begin{center}
    \textbf{
    \textit{How can we reduce communication overhead without resorting to time-consuming quantization or compressed sensing techniques, while still preserving the accuracy of the trained models?}
    }
  \end{center}
\end{mdframed}

\paragraph*{Our Contributions and Novelty}

In this study, we propose a communication-efficient over-the-air \underline{fed}erated learning via \underline{z}eroth-\underline{o}rder \underline{e}stimation (FedZOE). 
The methodology we provided has not been discussed in OtA-FL such as \cite{azimi2024over,yang2022over,yang2020federated,xu2022pruning,cao2020optimized}. 
Moreover, compared to existing ZOO-based FL methods~\cite{fang2022communication, malladi2023fine, 8364613} and LoRA-based FL approaches~\cite{hu2021lora, kuo2024federated}, our proposed framework achieves a significant reduction in communication overhead while maintaining convergence speed. This is accomplished by leveraging randomized gradient estimation (RGE) to efficiently estimate the local model update vectors, whereas conventional approaches use RGE to obtain the gradient vectors replacing the backpropagation step.

Motivated by the identified research challenges, the contributions of this work are multifaceted and can be summarized as follows:
\begin{itemize}
    \item We propose a novel OtA-FL framework that leverages RGE to compress the local model update vectors of EDs, enabling gradient-exchange-free OtA-FL while maintaining convergence speed.
    \item Utilizing shared random seeds, EDs transmit only a series of scalar values to the PS via OtA, allowing the aggregated gradient to be estimated using RGE. Notably, \textbf{8,192 symbols per single uplink transmission are sufficient to train a neural network with 11 million parameters} (\textbf{compression ratio $\approx$ 0.07\%}), as shown in \cref{fig:various_L_convergence}.   
    \item In the downlink phase, the proposed method allows the PS to transmit only the aggregated scalar values received from the EDs, significantly reducing communication overhead during the downlink.  
    \item We provide theoretical guarantees for the convergence of the proposed method.  
    \item Through extensive benchmarking on diverse scenarios, including datasets beyond Tiny-ImageNet, CIFAR-10, Brain-CT, SVHN, and CIFAR-100, and using ResNet models, we demonstrate that the proposed method effectively reduces communication overhead while achieving robust performance.  
\end{itemize}

\paragraph*{Notations}

The notations used in this study are as follows.
The sets of real numbers and complex numbers are denoted by $\mathbb{R}$ and $\mathbb{C}$, respectively.
Lowercase boldface letters represent column vectors, \eg, $\mathbf{h} \in \mathbb{C}^n$.
The real and imaginary parts of a complex number are denoted by $\mathtt{Re}\{\cdot\}$ and $\mathtt{Im}\{\cdot\}$, respectively.
The notation $|\mathcal{A}|$ denotes the cardinality of a set $\mathcal{A}$.
For an integer $K$, $[K]$ represents the set of integers from $1$ to $K$, \ie, $[K] = \{1, 2, \dots, K\}$.
$\mathbbm{1}_K \in \mathbb{R}^K$ denotes the vector of length $K$ whose entries are all ones.

\section{System Model}
\label{sec:system_model}

We consider an FL network environment comprising $K$ single-antenna EDs and a multi-antenna PS, \ie, an uplink multi-user single-input multi-output (UL-MU-SIMO) system. We assume that the PS is equipped with $N$ antennas, and the set of EDs is denoted by $[K] = \{1, \cdots, K\}$.

\paragraph*{Learning Modeling}

For EDs $k \in [K]$, we assume that each ED has a local dataset $\mathcal{D}_k$, which does not overlap with the datasets of other EDs, \ie, $\mathcal{D}_i \cap \mathcal{D}_j = \emptyset$ for all $i, j \in [K]$ and $i \neq j$.
For brevity of the notation, we assume that the local datasets for all EDs have the same size\footnote{We note that all the derivations in our work can be easily extended to the case where $|\mathcal{D}_k| \neq |\mathcal{D}_{k'}|$ for all $k,k'\in[K]$.} in the remainder of this paper, \ie, $|\mathcal{D}_k| = |\mathcal{D}_{k'}|$ for all $k,k'\in[K]$.
Consider a deep neural network (DNN) model parameterized by a vector $\mathbf{w}\in\mathbb{R}^S$, where $S$ denotes the total number of parameters. The local loss function of the $k$-th ED is given by
\begin{equation}\label{eq:local_loss}
    F_k(\mathbf{w}) = \frac{1}{|\mathcal{D}_k|} \sum_{d \in \mathcal{D}_k} \ell (\mathbf{w};d), 
\end{equation}
where $\ell$ denotes the loss function, \eg, mean square error or cross-entropy.
Aggregating the local loss functions from all EDs, the global loss function is defined as
\begin{equation}\label{eq:global_loss}
\begin{split}
    F(\mathbf{w}) & = \frac{1}{\sum_{k\in[K]}|\mathcal{D}_k|} \sum_{k\in[K]} |\mathcal{D}_k|F_k(\mathbf{w})\\ 
    & =\frac{1}{K} \sum_{k\in[K]} F_k(\mathbf{w}).\\ 
\end{split}
\end{equation}
The goal of FL is to minimize the global loss function in \eqref{eq:global_loss} by exchanging gradient vectors computed by EDs, instead of sharing the raw training data $\mathcal{D}_k$.

\paragraph*{Channel Modeling}

We adopt a block-fading channel model, where the uplink channel between ED $k$ and the PS is denoted by $\mathbf{h}_k\in\mathbb{C}^N$.
Representing the input signal of the $k$-th ED as $x_k$, the post-processed received signal at the PS is given by
\begin{equation}\label{eq:signal_model_original}
    y = \mathbf{r}^\mathrm{H}\left(\sum_{k\in[K]}\mathbf{h}_kx_k + \mathbf{n}\right),
\end{equation}
where $\mathbf{n}\in\mathbb{C}^{N}$ denotes the additive white Gaussian noise (AWGN) at the PS with the standard deviation of $N_0$, \ie, $\mathbf{n}\sim\mathcal{N}(0,N_0 \mathbf{I}_N)$, and $\mathbf{r}\in\mathbb{C}^N$ is the receive combiner.
In each ED, the average transmission power is limited by $P_\text{max}$, imposing the following constraint on the signal power:
\begin{equation}\label{eq:power_constraint}
    \mathbb{E}\left[ |x_k|^2 \right] \leq P.
\end{equation}

\begin{figure*}[t]
    \centering
    \includegraphics[width=0.99\linewidth]{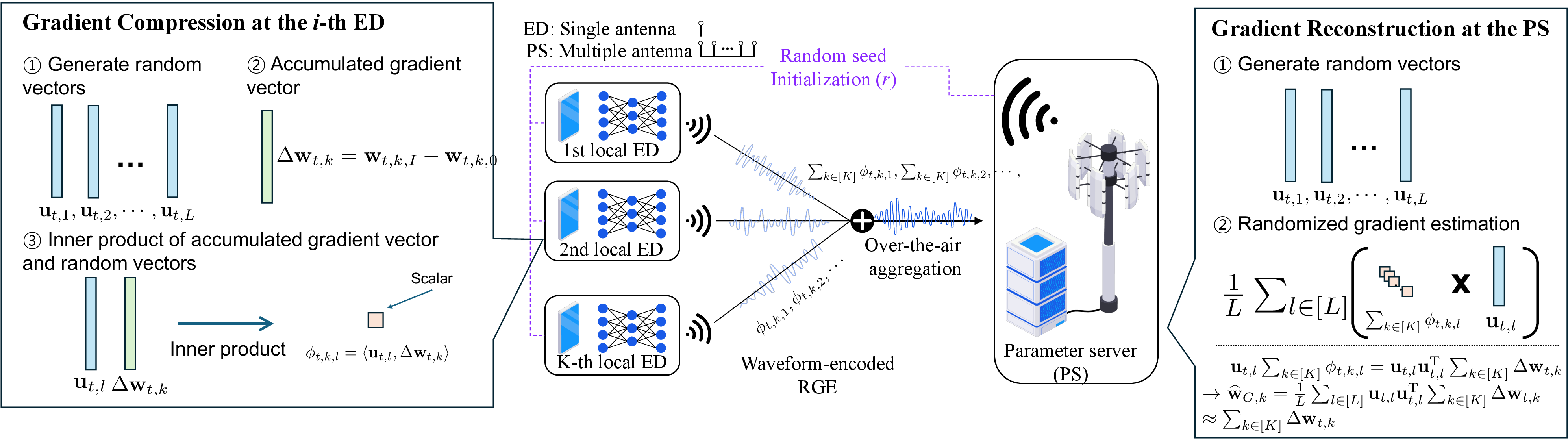}
    \caption{Overview of the proposed gradient compression and reconstruction framework in Fed-ZOE. The left panel illustrates the gradient compression process at each ED, where local model update vectors are projected onto randomized vectors, producing a series of scalar values of transmission. 
    The central penal depicts the aggregation of these scalar values across all EDs. The right panel shows the gradient reconstruction process at the PS via RGE. }
    \label{fig:Conceptual_diagram}
\end{figure*}

\section{Proposed Method}
\label{sec:proposed_method}

In this section, we present \textit{Fed-ZOE}, a method that leverages randomized gradient estimation (RGE) to reduce the communication overhead in OtA-FL. 
Unlike ZOO-related studies~\cite{fang2022communication, malladi2023fine, qin2023federated}, Fed-ZOE estimates local model update vectors rather than step-wise local gradients, achieving higher accuracy and computational efficiency. 
By doing so, we take only the advantages of the first-order and zeroth-order optimization. 
In the FL, the aim of the optimization is to find the following solution $\mathbf{w}_\text{opt}=\argmin_\mathbf{w} F(\mathbf{w})$ by the cooperation of the EDs without exchanging the data in the EDs. 
Denoting the total round of the FL as $T$; then, the $k$-th ED first updates its local model via $I$ local SGD steps, where the mini-batch is randomly sampled from the dataset $\mathcal{D}_k$. 
Similar to the conventional \texttt{FedAvg}, the proposed method consists of three steps: 1) local update, 2) global aggregation, and 3) weight broadcasting. 

As is customary in existing works~\cite{chellapandi2023federated,chu2022federated}, the proposed method starts with initializing random, but synchronized, weights in each ED. 
We denote the local parameter of ED $k$ in the $i$-th local update at communication round $t$ as $\mathbf{w}_{k,t,i}$. 
Representing the global model weights at the $t$-th communication round as $\mathbf{w}_{G,t}$, the initialized parameter vector at ED $k$ is $\mathbf{w}_{k, 0, 0} = \mathbf{w}_{G,0}$ for all $k\in[K]$.

\subsection{Local Update}

As per \texttt{FedAvg}, the EDs locally update DNN model weights for $I$ local update steps via stochastic gradient descent (SGD).
In each local training step, ED $k$ randomly samples a mini-batch $\mathcal{B}_{k,t,i} \subset \mathcal{D}_{k}$ with the size of $B$.
With the sampled mini-batch $\mathcal{B}_{k,t,i}$, the local weight $\mathbf{w}_{k,t,i}$ is updated by 
\begin{equation}
    \mathbf{w}_{k,t,i+1} = \mathbf{w}_{k,t,i}- \eta \mathbf{g}_{k,t,i}, \forall i\in[I],  
\end{equation}
where $\mathbf{g}_{k,t,i} = \frac{1}{B}\sum_{d\in\mathcal{B}_{k,t,i}}\nabla_\mathbf{w} F_k(\mathbf{w}_{k,t,i},d)$, and $\eta$ denotes learning rate.

\subsection{Local Model Update Vector Compression}

In each communication round, the EDs need to send the local model update vectors to the PS, where the local model update vector of the $k$-th ED at the $t$-th communication round is defined by 
\begin{equation}
\begin{split}
    \Delta \mathbf{w}_{k,t} & = \mathbf{w}_{k,t,I} - \mathbf{w}_{k,t,0} \\ 
                            & = -\eta\sum_{i\in[I]} \mathbf{g}_{k,t,i-1}, ~~ \forall k\in[K].
\end{split}
\end{equation}
In the OtA-FL, because one communication subcarrier is required to send one element of a local model update vector, the number of required communication subcarriers equals the number of parameters. 
Thus, as mentioned in \cref{sec:Introduction}, communication overheads are substantially heavy due to millions or billions of parameters.
To reduce the communication overhead in OtA-FL systems, we need to compress the local model update vectors while preserving the superposition property. That is, the aggregated gradient should be estimated by using the communication signals aggregated over the air. 

To this end, we leverage RGE for gradient compression and recovery. 
Before presenting the proposed method, we briefly review RGE~\cite{nesterov2017random}, which was originally designed to perform \textit{gradient-free} or \textit{back-propagation-free} optimization.

\begin{definition}[Randomized gradient estimation~\cite{nesterov2017random}]
\label{def:RGE}
    Consider the general optimization problem
    \begin{equation}
        \min_{\mathbf{w} \in \mathbb{R}^S} f(\mathbf{w}),    
    \end{equation}
    where $f:\mathbb{R}^S \rightarrow \mathbb{R}$ is the objective function.
    Then, we define the \textbf{smoothed} version of the objective function as 
    \begin{equation}
    f_\mu(\mathbf{w})  = \mathbb{E}_{\mathbf{u}}\left[ f(\mathbf{w} + \mu\mathbf{u})\right],
    \end{equation}
    where $\mu$ is an arbitrary constant, and $\mathbf{u}$ denotes an arbitrary spherically symmetric and unit-variance random noise.
    The gradient is then given by
    \begin{equation}
    \nabla f_\mu(\mathbf{w}) = \frac{\partial}{\partial \mathbf{w}} \mathbb{E}_{\mathbf{u}}[f(\mathbf{w} + \mu \mathbf{u})],
    \end{equation}
    and an estimated gradient is given by 
    \begin{equation}\label{eq:1p_zoo}
    \widehat{\nabla} f(\mathbf{w}) = \frac{f(\mathbf{w} + \mu \mathbf{u}) - f(\mathbf{w})}{\mu} \mathbf{u}.
    \end{equation}
\end{definition}

In \cref{def:RGE}, as $\mu\rightarrow 0$, the randomized gradient estimator yields $\widehat{\nabla} f(\mathbf{w}) \rightarrow \mathbf{u}^\mathrm{T} \nabla f(\mathbf{w})$, \ie, directional derivatives, thereby providing us an unbiased estimator of $\nabla f(\mathbf{w})$ as follows: $\mathbb{E}_\mathbf{u}\left[\mathbf{u}^\mathrm{T} \nabla f(\mathbf{w})\mathbf{u} \right] = \mathbb{E}_\mathbf{u}\left[\mathbf{u}\mathbf{u}^\mathrm{T} \nabla f(\mathbf{w}) \right] = \nabla f(\mathbf{w})$, which is reminiscent of random vector quantization in CSI feedback methods~\cite{Santipach_05,Au-yeung07,Santipach_09}.

\paragraph*{Compression and recovery strategy}

In our work, we bring the idea that $\mathbb{E}_{\mathbf{u}}\left[\mathbf{u}\mathbf{u}^\mathrm{T} \nabla f(\mathbf{w})\right] = \nabla f(\mathbf{w})$ to build our compression strategy. 
In \cref{fig:Conceptual_diagram}, we depict the overall system model and architecture of the proposed gradient compression and recovery.
For ED $k$, we have a local model update vector  $\Delta\mathbf{w}_{k,t}\in\mathbb{R}^S$, which will be compressed to reduce communication overhead. 
Our compression strategy is to compress the local model update vector with a series of random vectors $\mathbf{u}_{t,l}\in\mathbb{R}^S$, $l\in[L]$, where $L$ denotes the number of random vectors. 
Then, the local model update vector $\Delta\mathbf{w}_{k,t}$ is compressed and recovered by 
\begin{equation}\label{eq:compress_recover}
\begin{split}
\underbrace{\Delta\widehat{\mathbf{w}}_{k,t}}_{\text{Recovered}}
& \coloneqq \frac{1}{L}\sum_{l\in[L]} 
\underbrace{\mathbf{u}_{t,l}^T\Delta \mathbf{w}_{k,t}}_{= \phi_{t,k,l} }
\mathbf{u}_{t,l} 
\end{split}
\end{equation}
where $\phi_{t,k,l}$ is a compressed version of the local model update vector, and $\phi_{t,k,l}=\mathbf{u}_{t,l}^\mathrm{T}\mathbf{w}_{k,t}$.
As depicted in the left part of \cref{fig:Conceptual_diagram} the EDs only send $\phi_{t,k,l}$, $l\in[L]$, to the PS instead of sending the local model update vector $\Delta\mathbf{w}_{k,t}$.

\paragraph*{Compression ratio}
The compression ratio of the proposed method compared to conventional OtA-FL is $\frac{L}{S}$. 
As shown in numerical results, the proposed method has a comparable evaluation accuracy even with a compression ratio of 0.07\% $\left(\frac{7}{10000}\approx\frac{8192}{1.1\times 10^7}\right)$.

\paragraph*{Check: Superposition property}
Here, we verify that our proposed compression strategy satisfies the superposition property, a requirement for leveraging over-the-air aggregation. 
For brevity, we let the aggregated gradient at the $t$-th communication round by $\frac{1}{K}\sum_{k\in[K]}\mathbf{w}_{k,t}$. 
Then, because the recovered gradient in \eqref{eq:compress_recover} can be represented by a linear combination of random vectors $\mathbf{u}_{t,l}$, the superposition property holds for our proposed method, \ie,
\begin{equation}
    \frac{1}{L}\sum_{l\in[L]}\left(\frac{1}{K}\sum_{k\in[K]} \phi_{t,k,l}\right) \cdot \mathbf{u}_{t,l} = \frac{1}{K}\sum_{k\in[K]} \Delta \widehat{\mathbf{w}}_{k,t},
\end{equation}
where the expectation over $\mathbf{u}_{t,l}$ gives the actual gradient as 
\begin{equation}
\begin{split}
    &\mathbb{E}_\mathbf{u}\left[\frac{1}{L}\sum_{l\in[L]}\left(\frac{1}{K}\sum_{k\in[K]} \phi_{t,k,l}\right) \cdot \mathbf{u}_{t,l}\right] \\ 
    & = \frac{1}{K}\sum_{k\in[K]}\mathbb{E}_\mathbf{u}\left[\mathbf{u}\mathbf{u}^\mathrm{T}\right] \Delta\mathbf{w}_{k,t} = \frac{1}{K}\sum_{k\in[K]}\Delta\mathbf{w}_{k,t}.
\end{split}
\end{equation}

\subsection{Global Aggregation}

Once all the EDs finish their local update procedures, they send their compressed values $\boldsymbol{\phi}_{t,k}=[\phi_{t,k,1},\phi_{t,k,2},\cdots,\phi_{t,k,L}]^\mathrm{T}\in\mathbb{R}^L$ via over-the-air aggregation.
Before the transmission phase, each of the EDs normalizes the compressed values by zero-mean and unit-variance, \ie, $\bar{\boldsymbol{\phi}}_{k,t} = \frac{\boldsymbol{\phi}_{k,t} - \mu_{k,t}}{\sigma_{k,t}}$, where $\mu_{k,t} = \frac{1}{L}\sum_{l\in[L]}\phi_{k,t,l}$ and $\sigma_{k,t}^2 = \frac{1}{L}\sum_{l\in[L]}\left(\phi_{k,t,l}-\mu_{k,t}\right)^2$.
We assume that $\mu_{k,t}$ and $\sigma_{k,t}$ are known at the PS because exchanging two scalar values requires negligible communication overhead.
Then, the received signal at the PS is formulated by 
\begin{equation}\label{eq:received}
    \mathbf{y}_{\text{G}}^\mathrm{H} = \mathbf{r}^\mathrm{H}\left(\sum_{k=1}^{K} \mathbf{h}_k b_k \bar{\boldsymbol{\phi}}_{k,t}^\mathrm{H} + \mathbf{n} \mathds{1}_{S}^\mathrm{T} \right) + \underbrace{\frac{\mathbf{r}^\mathrm{H}}{K}\sum_{k=1}^{K}\mathbf{h}_k b_k\frac{\mu_{k,t}}{\sigma_{k,t}}}_{(a)},
\end{equation}
where the power amplification $b_k$ is constrained by $\mathbb{E}[|b_k|^2]<P$ to satisfy the constraint \eqref{eq:power_constraint}.
We note that the additive term $(a)$ in \eqref{eq:received} compensates the shifted value of $\boldsymbol{\phi}_{k,t}$ in the normalization, which is known at the PS.
Because the aim of the global aggregation is to estimate the average of compressed values $\boldsymbol{\phi}_{k,t}$, we can formulate an optimization as
\begin{equation}\label{eq:MSE_biased}
    \min_{b_k, \mathbf{r}} \mathbb{E}\left[\left\Vert \mathbf{y}_\text{G} - \frac{1}{K}\sum_{k=1}^K \boldsymbol{\phi}_{k,t}\right\Vert^2\right], 
\end{equation}
which can be rewritten by 
\begin{problem} \label{eq:P1}
    \begin{alignat}{3}
        \text{\ref{eq:P1}: } & & \min_{\mathbf{r}, b_k} ~~ & \underbrace{\left\Vert\mathbf{r}^\mathrm{H}\left(\sum_{k=1}^{K} \left(\frac{\mathbf{h}_k b_k}{\sigma_{k,t}} - \frac{1}{K}\right) \boldsymbol{\phi}_{k,t}\right) \right\Vert^2}_{\text{Bias}}+ \underbrace{S\Vert\mathbf{r}\Vert_2^2}_{\text{Variance}} \label{P1:obj} \nonumber
    \end{alignat}
\end{problem}
In Problem \ref{eq:P1}, the solution of $b_k$ for given $\mathbf{r}$ can be denoted as 
\begin{equation}\label{eq:sol_bk}
        b_k = \frac{\mathbf{h}_k^\mathrm{H}\mathbf{r}}{|\mathbf{h}_k^\mathrm{H}\mathbf{r}|}\cdot\min\left(\sqrt{P}, \frac{\sigma_{k,t}}{K|\mathbf{r}^\mathrm{H}\mathbf{h}_k|} \right), \forall k\in[K].
\end{equation}

\paragraph*{Constraint for unbiased estimator}

As shown in \cite{ajalloeian2020convergence}, the biased gradient estimation results in slowed convergence or diverge of ML training algorithms.
Hence, we aim to remove bias in the estimated gradient by scaling $b_k$ and $\mathbf{r}$ to satisfy $\mathbb{E}[\mathbf{y}_{\text{G}}]=\frac{1}{K}\sum_{k\in[K]}\boldsymbol{\phi}_{k,t}$.
By substituting \eqref{eq:sol_bk} into the objective function of Problem \ref{eq:P1}, we confirm that $\mathbb{E}[\mathbf{y}_{\text{G}}]=\frac{1}{K}\sum_{k\in[K]}\boldsymbol{\phi}_{k,t}$ if $b_k$ in \eqref{eq:sol_bk} satisfies $b_k=\frac{\mathbf{r}^\mathrm{H}\mathbf{h}_k\sigma_{k,t}}{K|\mathbf{r}^\mathrm{H}\mathbf{h}_k|^2}$.
That is, the gradient estimation is unbiased if 
\begin{equation}\label{eq:constraint_unbiased}
    |\mathbf{r}^\mathrm{H} \mathbf{h}_k| \ge \frac{\sigma_{k,t}}{K\sqrt{P}}.
\end{equation}


\paragraph*{Beamforming problem formulation}
With the constraint \eqref{eq:constraint_unbiased}, the objective function of Problem \ref{eq:P1} is represented by $S\Vert\mathbf{r}\Vert^2$.
Then, the problem is formulated by uniform-forcing over-the-air beamforming problem, which is previously discussed in \cite{8364613}:
\begin{problem} \label{eq:P2}
    \begin{alignat}{3}
        \text{\ref{eq:P2}: } & & \min_{\mathbf{r}} ~~ & \Vert\mathbf{r}\Vert_2^2 \label{P2:obj} & \\
                         {}  & & ~\text{s.t. } ~~ & |\mathbf{r}^\mathrm{H}\mathbf{h}_k| \ge \frac{\sigma_k}{K\sqrt{P}},~~ \forall k\in[K]. &\label{eq:P2_constraint}
    \end{alignat}
\end{problem}

For brevity of the main part of this paper, the solution of Problem \ref{eq:P2} is analyzed in Appendix \ref{sec:appendix_beamforming}.
In the remainder part, we denote that the receive combiner solution $\mathbf{r}$ of Problem \ref{eq:P2} as $\mathbf{r}_\text{opt}$. 


\paragraph*{Aggregation}

With the receive combiner $\mathbf{r}_\text{opt}$, we have 
\begin{align}
        \mathbf{y}_{\text{G}}^\mathrm{H} & = \mathbf{r}_\text{opt}^\mathrm{H}\left(\sum_{k=1}^{K} \mathbf{h}_k b_k \bar{\boldsymbol{\phi}}_{k,t}^\mathrm{H} + \mathbf{n} \mathds{1}_{S}^\mathrm{T} \right) + \frac{\mathbf{r}_\text{opt}^\mathrm{H}}{K}\sum_{k=1}^{K}\mathbf{h}_k b_k\frac{\mu_{k,t}}{\sigma_{k,t}}, \notag\\ 
        & = \frac{1}{K}\sum_{k=1}^{K}\boldsymbol{\phi}_{k,t}^\mathrm{H}  + \mathbf{r}_\text{opt}^\mathrm{H} \mathbf{n} \mathds{1}_{S}^\mathrm{T} \in \mathbb{R}^{L}.   \label{eq:proposed_aggr}
\end{align}
Then, from the received signal $\mathbf{y}_{\text{G}}$, the global model can be updated with the recovered gradient given by:
\begin{equation}\label{eq:def_delta_w}
\begin{split}
    \Delta\widehat{\mathbf{w}}_{G,t} & = \frac{1}{L} \sum_{l\in[L]} \mathbf{u}_{t,l}[\mathbf{y}_{\text{G}}]_l \\ 
    & = \frac{1}{L} 
    \underbrace{\begin{bmatrix}
           \mathbf{u}_{t,1} & \mathbf{u}_{t,2} & \cdots & \mathbf{u}_{t,L}
   \end{bmatrix}}_{=\mathbf{U}_t\in\mathbb{R}^{S\times L}}
   \mathbf{y}_{\text{G}}.
\end{split}
\end{equation}

\subsection{Downlink Broadcast}

After the aggregation step in \eqref{eq:proposed_aggr}, the parameter server broadcasts the updated weights to EDs. 
In conventional OtA-FL methods, the downlink broadcast communication also causes substantial communication overhead.

In our approach, we leverage the shared random seed to send only 
$\frac{1}{L}\mathbf{y}_G = \frac{1}{K}\sum_{k\in[K]}\widehat{\boldsymbol{\phi}}_{k,t}$, which requires significantly less communication overhead compared to sending overall network parameter changes $\Delta\widehat{\mathbf{w}}_t$. 
In \cref{alg:FEDZOE}, we show the step-by-step details of the proposed Fed-ZOE method.




\begin{algorithm}[!t]
\caption{\texttt{Fed-ZOE} Algorithm.}
\label{alg:FEDZOE}
\begin{algorithmic}[1] 
\STATE \textbf{Initialize}: Total communication round $T$, total local steps $I$, number of EDs $K$, learning rate $\eta$, number of RGE samples $L$, and weights $\mathbf{w}_{G,0}$ with dimension $S$, random seed $r$. 
\STATE Set random seed $r$ for all EDs $k\in[K]$.
\FOR {$t=0$ to $T-1$}
    \STATE \underline{\textbf{ON CLIENTs (Local training and compression):}}
    \FORALLP {\textbf{client $k\in[K]$}} \Comment{Local training}
        \STATE $\mathbf{w}_{t,k,0} \leftarrow \mathbf{w}_{G,t}$. 
        \FOR {$i=1$ to $I$}
            \STATE $\mathbf{w}_{k,t,i+1} \leftarrow \mathbf{w}_{k,t,i}- \frac{\eta}{B}\sum_{d\in\mathcal{B}_{k,t,i}}\nabla_\mathbf{w} F_k(\mathbf{w}_{k,t,i},d)$.
        \ENDFOR
        \STATE $\Delta \mathbf{w}_{k,t} \leftarrow \mathbf{w}_{k,t,I} - \mathbf{w}_{k,t,0}$.
    \ENDFAP
    \FORALLP {\textbf{client $k\in[K]$}} \Comment{Compression step}
        \STATE $\mathbf{U}_t \in\mathbb{R}^{S\times L}$  from seed $r$. \Comment{RGE samples}
        \STATE $\boldsymbol{\phi}_{k,t} \leftarrow \mathbf{U}_{t}^\mathrm{T} \Delta\mathbf{w}_{k,t}$. \Comment{Compression step}
        \STATE $\bar{\boldsymbol{\phi}}_{k,t} \leftarrow \frac{\boldsymbol{\phi}_{k,t} - \mu_{k,t}}{\sigma_{k,t}}$.
        \STATE Send $\sigma_k$ to the PS. 
    \ENDFAP 
    \vspace{4pt} \STATE \underline{\textbf{ON the PS (Beamforming optimization):}}
    \STATE Compute $\mathbf{r}$ by solving Problem \ref{eq:P2}. 
    \STATE Broadcast $\mathbf{r}$.  
    \vspace{4pt} \STATE \underline{\textbf{ON CLIENTs (Transmission):}} 
    \FORALLP {\textbf{client $k\in[K]$}} \Comment{Aggregation}
        \STATE Send $\bar{\boldsymbol{\phi}}_{k,t}$ to the PS via over-the-air aggregation. 
    \ENDFAP 
    \vspace{4pt} \STATE \underline{\textbf{ON the PS (Aggregation and broadcasting):}}
    \STATE Broadcast $\frac{1}{L}\mathbf{y}_G\in\mathbb{R}^{L}$ to the EDs $[K]$.  

    \vspace{4pt} \STATE \underline{\textbf{ON CLIENTs (Update with aggregated gradient):}} 
    \FORALLP {\textbf{client $k\in[K]$}} \Comment{Local training}
    \STATE $\Delta\widehat{\mathbf{w}}_{G,t} \leftarrow \frac{1}{L}\mathbf{U}_{t}\mathbf{y}_\mathrm{G}$.
    \STATE $\mathbf{w}_{G,t+1} \leftarrow \mathbf{w}_{G,t} + \Delta
            \widehat{\mathbf{w}}_{G,t}$. 
    \ENDFAP
    
\ENDFOR

\STATE \textbf{Output}: Trained neural network weight $\mathbf{w}_{G,T}$.
\end{algorithmic}
\end{algorithm}

\section{Theoretical Results}
\label{sec:Convergence}

In this section, we show the convergence guarantee of the proposed method. 
Before starting the proof of the convergence, we introduce the assumptions required in the proof. 
Similar to the previous studies, we have the following assumption.
\begin{assumption}
\label{assumption:1}
The local loss function $F_k$ and the global loss function $F$ have the following assumptions:
\begin{enumerate}[label=(A\arabic*), leftmargin=3em]
    \item The global loss function $F(\mathbf{w})$ is finite and lower-bounded, \ie, $F(\mathbf{w}) \ge \min_\mathbf{w} F(\mathbf{w}) > -\infty$ for all $\mathbf{w}\in\mathbb{R}^S$.
    \item For all $k\in[K]$, $\mathbf{w}$, and $\mathbf{w}'$,  the local loss function $F_k(\mathbf{w})$ satisfies $
        \Vert \nabla F_k(\mathbf{w}') - \nabla F_k(\mathbf{w}) \Vert \le \beta \Vert \mathbf{w}' - \mathbf{w} \Vert$.
    \item The mini-batch gradient $\mathbf{g}_{k,t,i}$ is unbiased and has bounded variance, \ie, $\mathbb{E}[\mathbf{g}_{t,k,i}]=\nabla F_k(\mathbf{w}_{k,t,i})$ and $\mathbb{E}[\Vert\mathbf{g}_{k,t,i} - \nabla F_k (\mathbf{w}_{k,t,i})\Vert^2] \le \xi^2$.
    \item For a constant $G_1$, we have $\Vert \nabla F_k(\mathbf{w})\Vert^2 \le G_1^2$.
\end{enumerate}
\end{assumption}

\begin{lemma}
    From the assumptions (A3) and (A4) in Assumption \ref{assumption:1}, we have 
    \begin{equation}
        \mathbb{E}[\Vert\mathbf{g}_{k,t,i}\Vert^2] \le \xi^2 + \Vert\nabla F_k(\mathbf{w}_{k,t,i})\Vert^2  \le \xi^2 + G_1^2 = G_2^2.
    \end{equation}
\end{lemma}

\begin{proposition}[]
\label{prop:proposition_1}
    Suppose that the assumptions in Assumption \ref{assumption:1} hold. The mean and variance of the recovered gradient $\frac{1}{K}\sum_{k\in[K]}\Delta\widehat{\mathbf{w}}_{k,t}$ are 
    \begin{equation}
        \begin{cases}
            \mathbb{E}\left[\frac{1}{K}\sum_{k\in[K]}\Delta\widehat{\mathbf{w}}_{k,t} \right]=- \frac{\eta}{K}\sum_{k\in[K]}\sum_{i\in[I]} \underbrace{\mathbb{E}[\mathbf{g}_{k,t,i-1}]}_{=\nabla F_k(\mathbf{w}_{k,t,i-1})} \\ 
            \mathtt{Var}\left[\frac{1}{K}\sum_{k\in[K]}\Delta\widehat{\mathbf{w}}_{k,t} \right] = \underbrace{\frac{I^2 S N_0 \eta_t^2 \Vert\mathbf{r}_\mathbbm{1}\Vert^2 G_2^2}{L^2}}_{\text{Channel Noise}} \\ 
            ~~~~~~~~~~~~~~~~~ + \underbrace{\frac{1}{K^2L}\sum_{k\in[K]} (3S-1)\Vert\Delta\mathbf{w}_{k,t}\Vert^2}_{\text{Compression Loss}},
        \end{cases}
    \end{equation}
    where $\mathbf{r}_\mathbbm{1}$ denotes the solution of Problem \ref{eq:P2} when $\sigma_{k,t} = 1$ for all $k\in[K]$. 
        The proof can be found in Appendix \ref{sec:appendix_proposition1}.
\end{proposition}
In \cref{prop:proposition_1}, we obtain the mean and variance of the recovered gradient $\Delta\widehat{\mathbf{w}}_{G,t}$. 


\begin{theorem}[Convergence]
\label{thm:convergence}
    Suppose that the assumptions in Assumption \ref{assumption:1} hold. If we choose a learning rate of $\eta_t=\frac{1}{\sqrt{TI}}$ with sufficient communication rounds and local steps satisfying $\frac{1}{\sqrt{TI}} \le \frac{2}{\beta\left(1+\frac{(3S-1)I}{K^2L}\right)} $, we have
    \begin{equation}
        \begin{split}
            &\frac{1}{TI}\sum_{t\in[T]}\sum_{i\in[I]}  \mathbb{E}\left\Vert \nabla F(\bar{\mathbf{w}}_{G,t-1,i-1})\Vert^2\right] \\
            &\le \frac{1}{\sqrt{T}}\bigg(\frac{2}{\sqrt{I}}\mathbb{E}\left[F(\mathbf{w}_{G,0}) - F(\mathbf{w}_{G,T})\right] + \frac{\xi^2\beta}{2KI^{\frac{3}{2}}} \\
            & ~~~~~~~~~~ + \frac{\beta \sqrt{I} S N_0  \Vert\mathbf{r}_\mathbbm{1}\Vert^2 G_2^2}{L^2} +  \frac{ 4 \beta^2 (I-1) G_1^2}{3\sqrt{T}K}\bigg) . 
        \end{split}
    \end{equation}
        The proof can be found in Appendix \ref{appendix:proof_convergence}.
\end{theorem}
In \cref{thm:convergence}, we show that by choosing a suitable learning rate, the algorithm's expected average gradient norm decreases at the rate of $\sqrt{1/T}$, similar to classic stochastic optimization. 
Interestingly, we require more samples $L$ for a higher learning rate $\alpha$, which will be demonstrated in \cref{fig:lr_samples}.

\section{Numerical Results}
\label{sec:numerical_results}

In this section, we demonstrate numerical results for in-depth insights into the proposed method (Fed-ZOE) under OtA-FL settings. 

\subsection{Experimental Setup}

In our experiments, we consider a wireless network with one PS and $K\in\{20,40,80,160\}$ EDs. 
The configuration of the channel coefficients is based on 3GPP standards~\cite{3gpp.38.211,3gpp38901}.
The maximum transmission power of EDs $P$ is 23 dBm, the number of antennas at the PS is 8, the additive white Gaussian noise power is -174 dBm/Hz, and the outdoor-to-indoor (O2I) channel model is used. 
We note that all the PS and EDs are distributed uniformly in a network with a 500-meter radius.
In each communication round, 10 users randomly participate in the training procedure. 

\begin{figure}[t]
    \centering
    \subfloat[Test accuracy versus communication rounds.\label{fig:convergence_cifar10_a}]{\includegraphics[width = 0.9\linewidth]{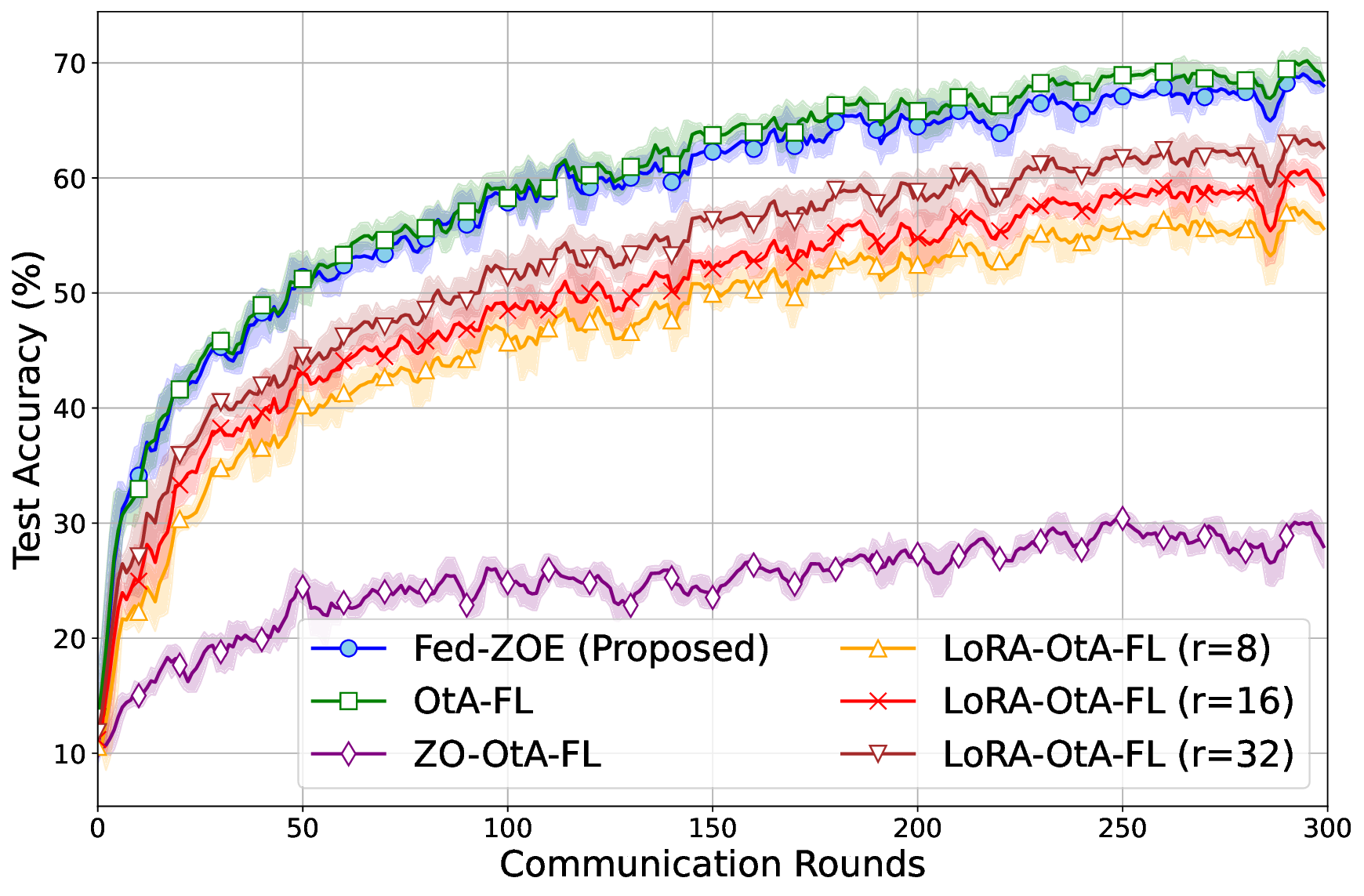}} \\    
    \subfloat[Test accuracy versus total communication and computation loads.\label{fig:convergence_cifar10_b}]{\includegraphics[width = 0.9\linewidth]{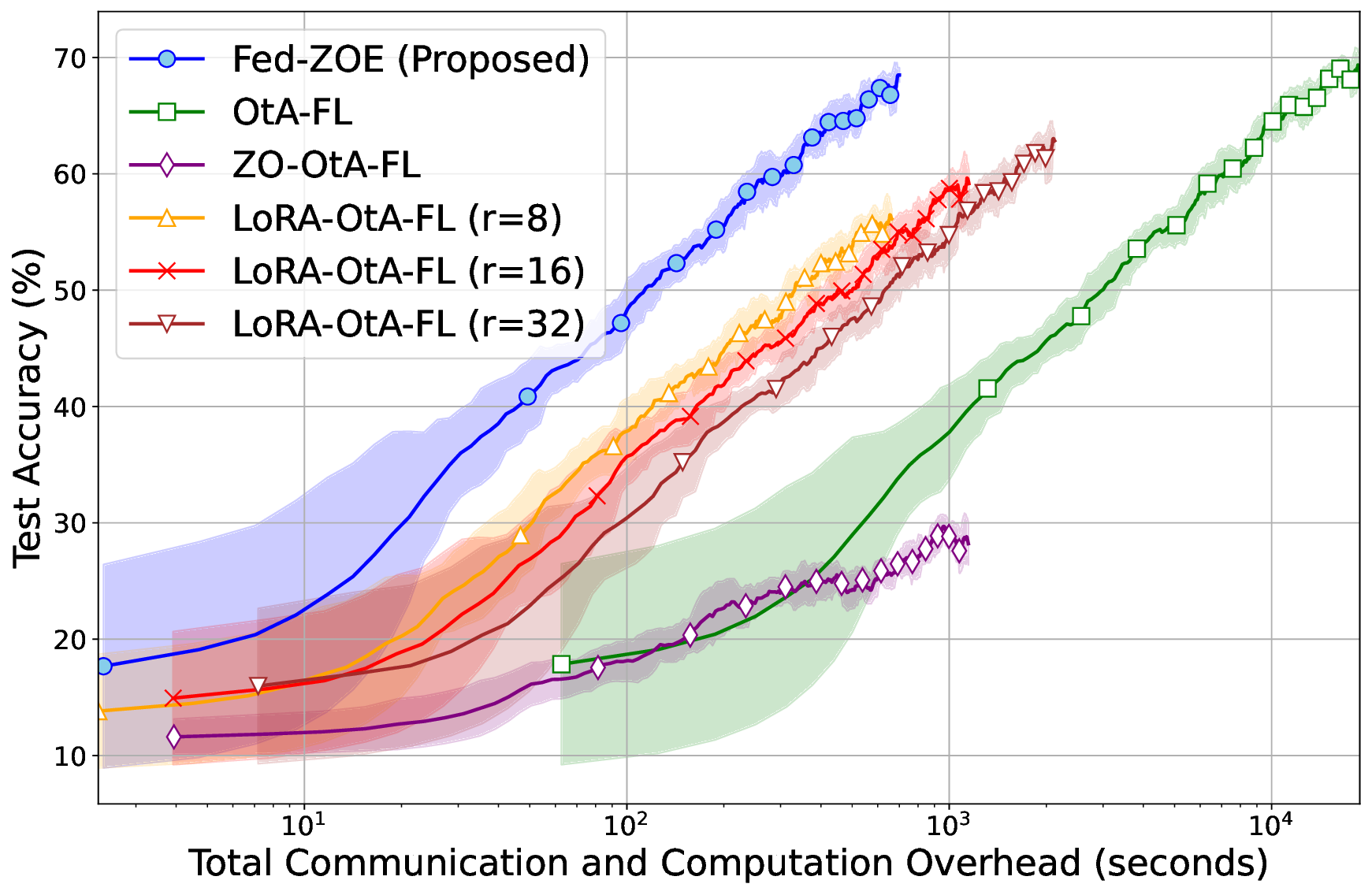}}
    \caption{
    Convergence evaluation of the proposed method and baseline methods. For this experiment, we train a ResNet-18 model with the CIFAR-10 dataset. 
    In \protect\subref{fig:convergence_cifar10_a}, the test accuracy for the communication rounds is depicted. 
    On the other hand, in \protect\subref{fig:convergence_cifar10_b}, the test accuracy for the total communication and computation loads is depicted. 
    }
    \label{fig:convergence_cifar10}
\end{figure}

\paragraph*{Datasets}

Five different datasets are used to verify the generalized advantages our method: 1) CIFAR-10~\cite{krizhevsky2009learning}, 2) CIFAR-100, 3) SVHN~\cite{netzer2011reading}, 4) Brain-CT~\cite{msoud_nickparvar_2021}, and 5) Tiny-ImageNet~\cite{tiny-imagenet}. 
For the data partitioning, we use the Dirichlet distribution, which is the standard method for simulating non-IID data in FL scenarios. 
By varying the parameter $\alpha$ of the Dirichlet we can control how data is distributed across EDs by generating probabilities for each class within a client's dataset: smaller $\alpha$ values create highly skewed distribution, while larger $\alpha$ values lead to more balanced, IID-like distribution.
In this work, we set the value of $\alpha$ as 0.5 for all experiments. 

\paragraph*{Learning configuration}

In the experiments, we configure the total communication round as $T=300$ for all datasets except the Brain-CT dataset. For the Brain-CT dataset, the number of communication rounds is $T=80$. 
In the local training, each ED trains $I\in\{20, 40, 80\}$ local training steps, where the batch size is configured by 64. 
Also, we choose ResNet-18~\cite{he2016deep} model and CIFAR-10, Tiny-ImageNet, and SVHN, CIFAR-100, and Brain-CT datasets.
Also, the number of sampled random noise $L$ is chosen in a set $\{512, 1024, 2048, 4096, 8192, 16384\}$, where the default value is 8,192.

\paragraph*{Baselines}

For a comprehensive comparison with the proposed method, we implement the following baseline schemes:
\begin{itemize}
    \item \textbf{OtA-FL~\cite{yang2020federated}:} A foundational method that uses FedAvg algorithm with over-the-air aggregation. All the hyper-parameters are the same with our method. 
    \item \textbf{LoRA-OtA-FL~\cite{oh2024communication}:} The low-rank adaptation (LoRA), a parameter-efficient fine-tuning method, can be used to reduce the communication overhead, as it can reduce the number of trainable parameter by adding low-rank adapters into CNN or fully-connected layers. In our experiments, we use two options of ranks $r$ and scaling parameters $\alpha_{\text{LoRA}}$: $(r,\alpha)\in\{(8,16), (16,32), (32,64)\}$.
    \item \textbf{Fed-ZO~\cite{qin2023federated,mhanna2024rendering,neto2024communication}:} The zeroth-order optimization (ZOO) also can be leveraged to reduce the communication overhead in OtA-FL systems. 
    As par~\cite{qin2023federated}, the ZOO-based approaches generally require more local training steps, we use the local training step of 200. 
\end{itemize}
We note that the advanced OtA-FL methods related to reducing communication overhead are addressed. 

\subsection{Experimental Results}

\begin{figure}
    \centering
    \includegraphics[width=0.99\linewidth]{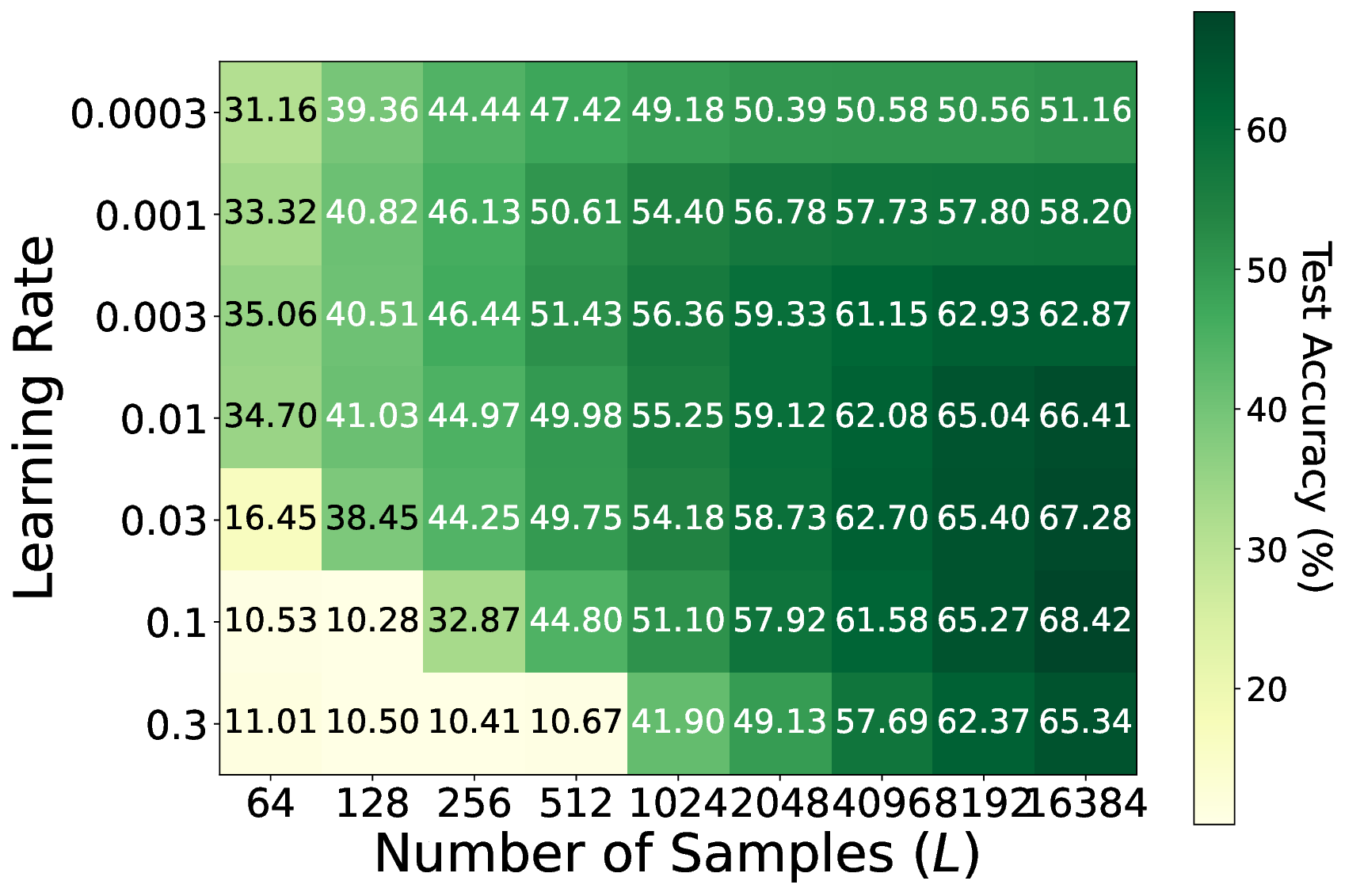}
    \caption{Test accuracy of the trained ResNet-18 models with various values of learning rate $\alpha$ and the number of RGE samples $L$. In this experiment, the proposed method is only implemented with $T=200$. }
    \label{fig:lr_samples}
\end{figure}

\begin{figure}
    \centering
    \includegraphics[width=1.0\linewidth]{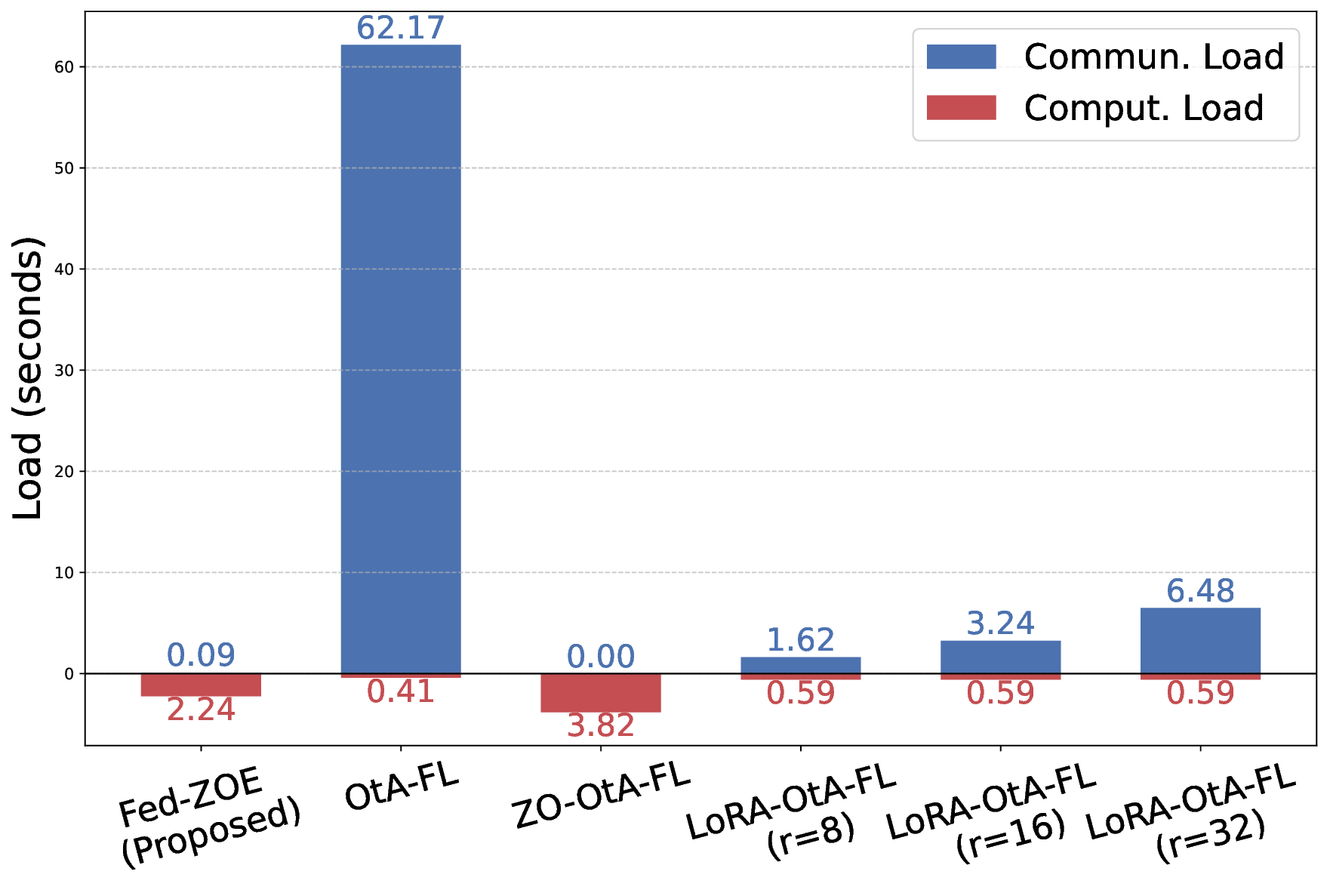}
    \caption{Communication and computation loads of the FL methods. The CIFAR-10 dataset and ResNet-18 model are used in this experiment. The unit is seconds.}
    \label{fig:loads_comparison}
\end{figure}

In this subsection, we execute experiments on ResNet-18 models with the CIFAR-10 dataset. In our numerical results, we depict the convergence of the FL methods with a moving average of the test accuracy due to fluctuating test accuracy results, where shaded regions denote the moving variance of the test accuracy. 

\paragraph*{Convergence Analysis}

In \cref{fig:convergence_cifar10}, the convergence of the proposed method and the baseline methods are depicted. 
First, in \cref{fig:convergence_cifar10_a}, the test accuracy is evaluated for each communication round. Compared to the LoRA-OtA-FL and ZO-OtA-FL methods, the proposed method outperforms the baseline methods by performing more efficient gradient compression via RGE-based compression.
Also, the proposed method closely achieves the test accuracy of the OtA-FL, which is a non-compressive method, thereby causing substantial communication overhead. 
In order to verify the efficiency of the FL approaches, we show the test accuracy versus the total communication and computation loads in \cref{fig:convergence_cifar10_b}. 
In the figure, the proposed method is the most efficient approach among the FL methods by virtue of its meager compression ratio.

For more in-depth discussions of the proposed method's convergence, we depict the test accuracy of the trained models with various values of learning rate $\alpha$ and the number of RGE samples $L$ in \cref{fig:lr_samples}.
As depicted in the figure, for higher learning rate $\alpha$, we have a higher threshold on $L$ for convergence.
This result perfectly matches the converge condition in \cref{thm:convergence}, which yields that a smaller learning rate $\alpha$ is required for a small number of samples $L$.


\begin{figure}
    \centering
    \includegraphics[width=1.0\linewidth]{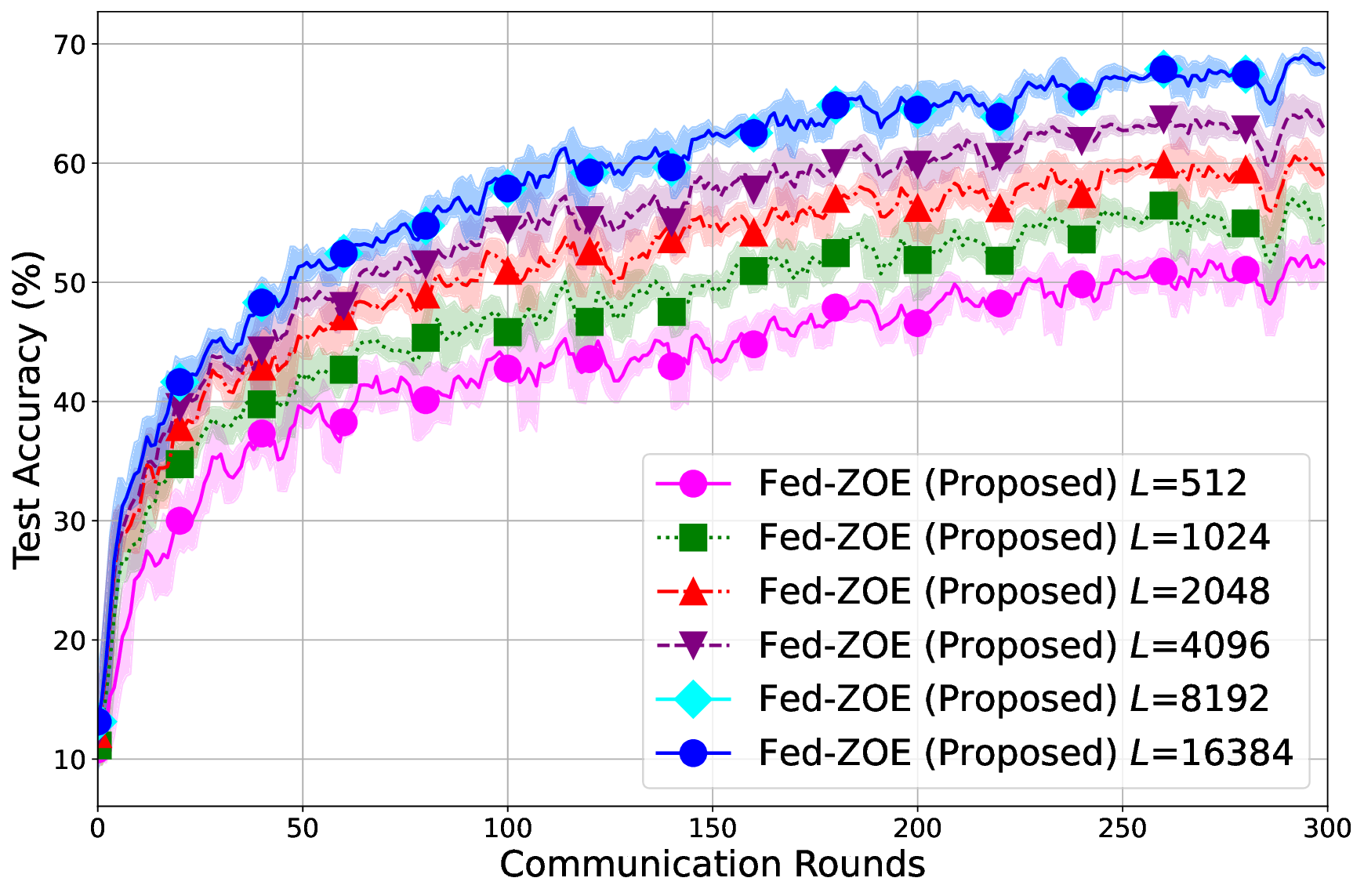}
    \caption{Convergence of the proposed method for various numbers of the RGE samples $L\in\{512, 1024, 2048, 4096, 8192, 16384\}$. }
    \label{fig:various_L_convergence}
\end{figure}

\begin{figure}
    \centering
    \includegraphics[width=1.0\linewidth]{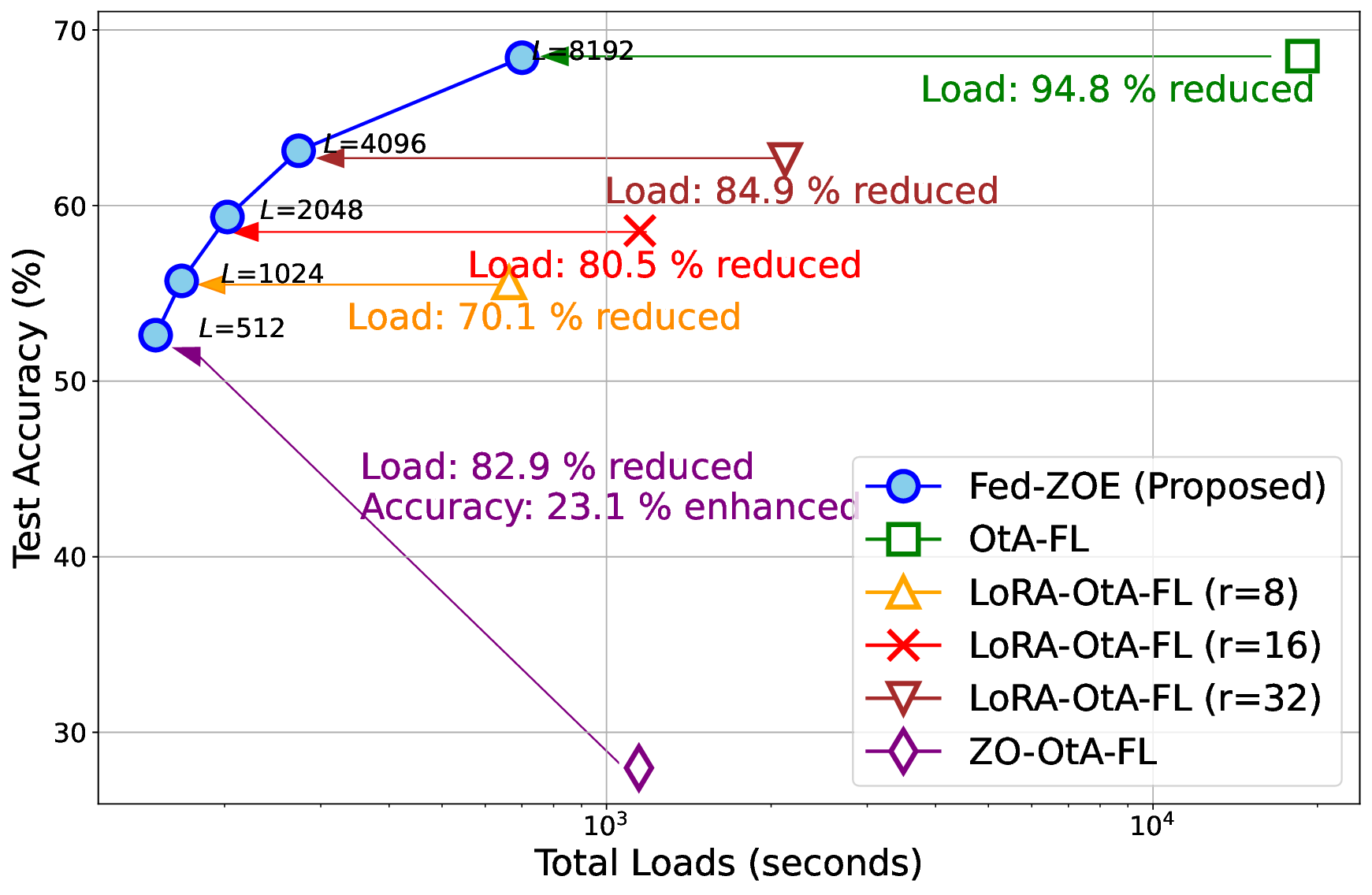}
    \caption{Tradeoff between the communication/computation overhead and the test accuracy for various numbers of RGE samples $L\in\{512, 1024, 2048, 4096, 8192\}$. In this experiment, we train a ResNet-18 model for the CIFAR-10 dataset.}
    \label{fig:various_samples}
\end{figure}

\begin{figure}
    \centering
    \subfloat[Test accuracy versus communication rounds ($K=20$).\label{subfig:various_EDs_a}]{\includegraphics[width=0.49\linewidth]{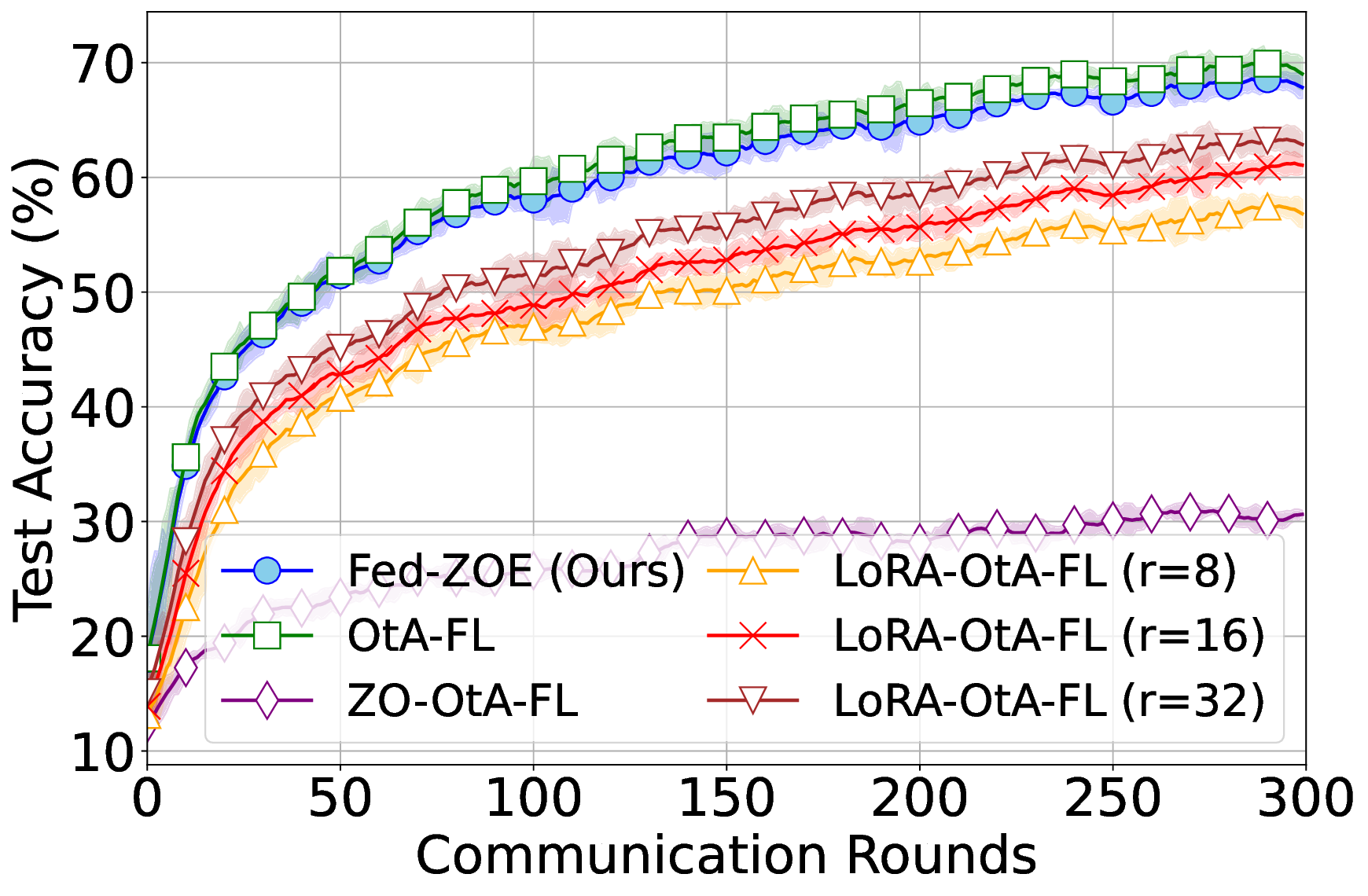}}
    \hfill
    \subfloat[Test accuracy versus communication/computation overheads ($K=20$). \label{subfig:various_EDs_b}]{\includegraphics[width=0.49\linewidth]{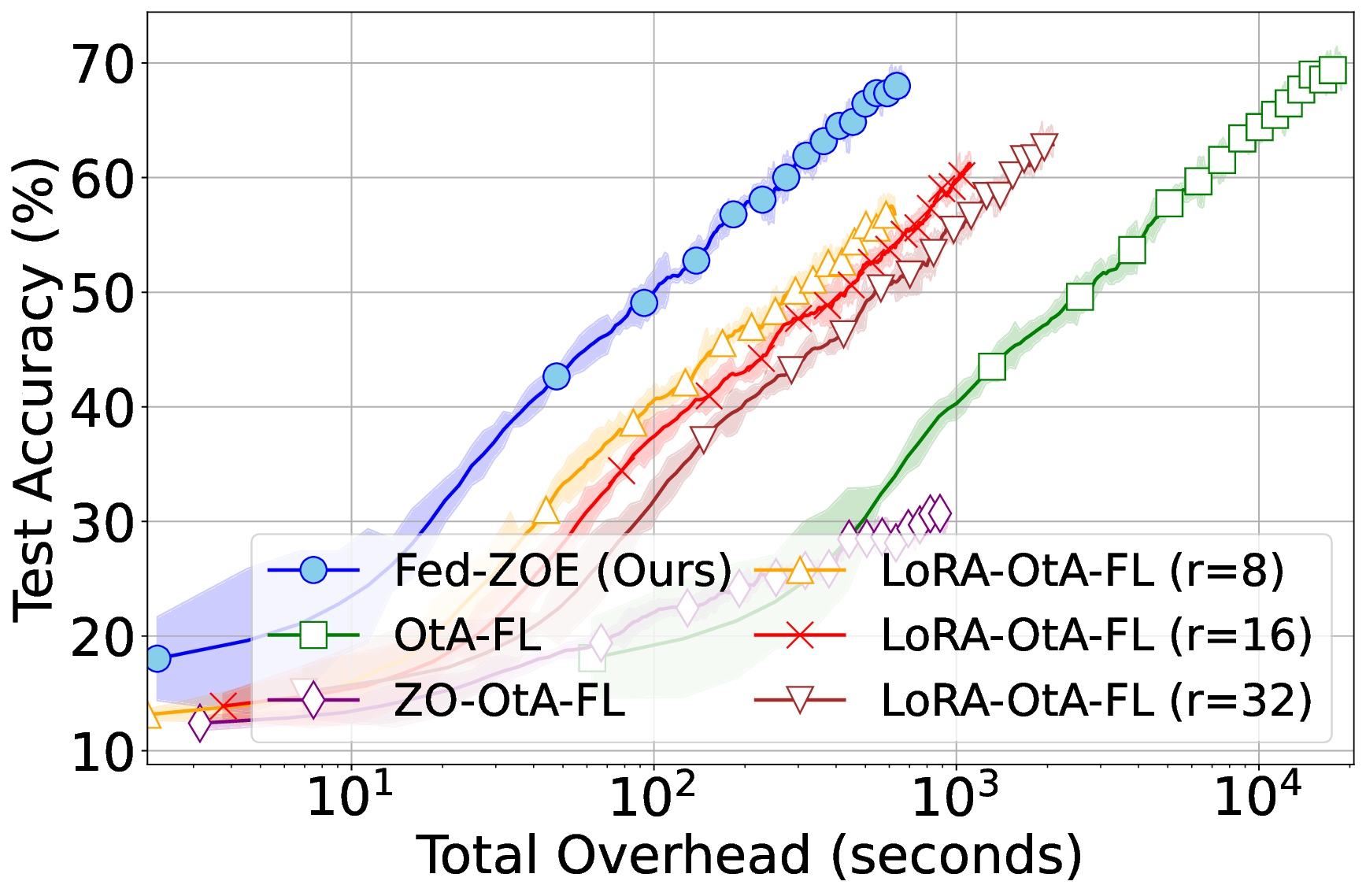}}\\
    \subfloat[Test accuracy versus communication rounds ($K=80$).\label{subfig:various_EDs_c}]{\includegraphics[width=0.49\linewidth]{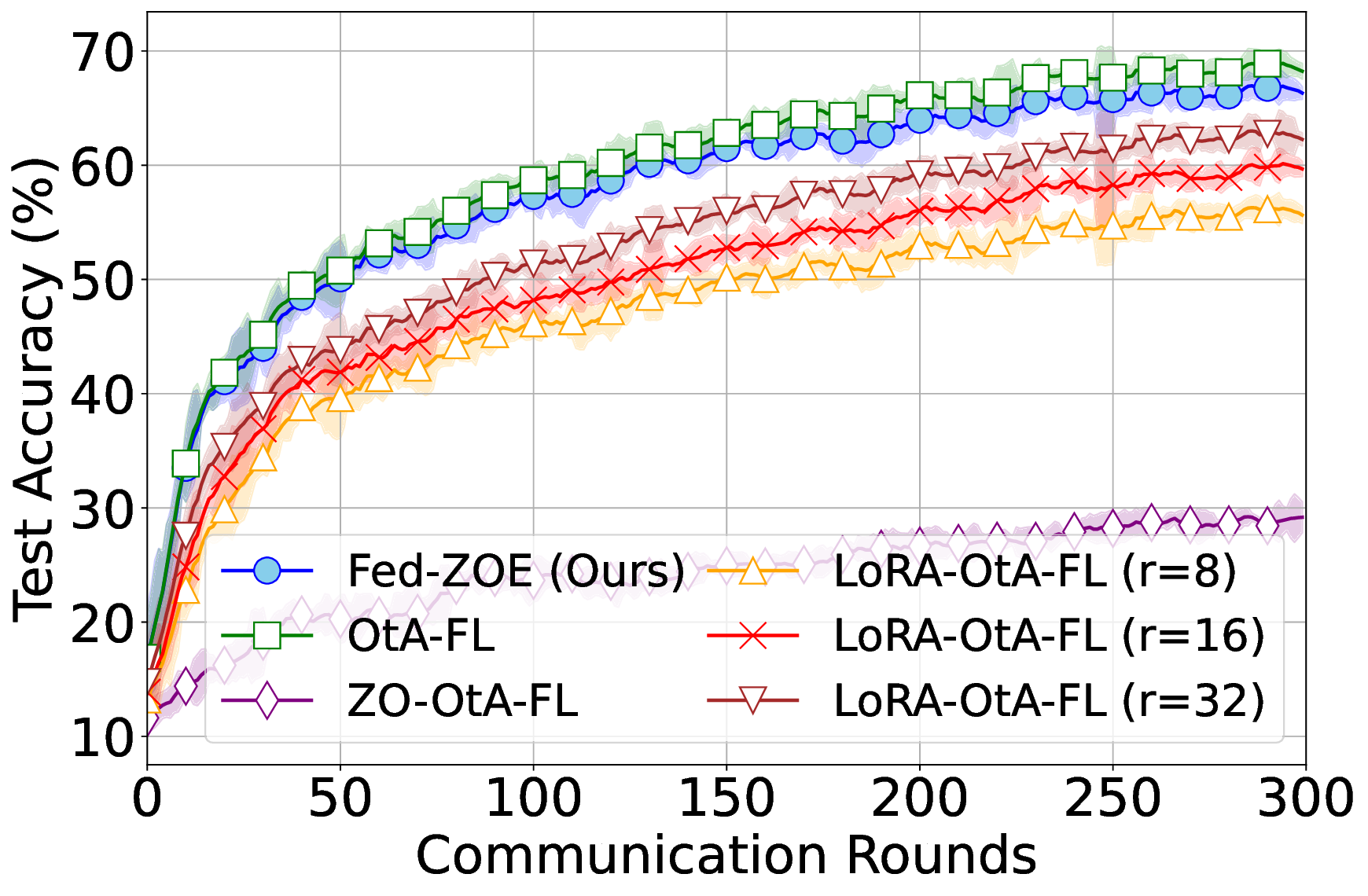}}
    \hfill
    \subfloat[Test accuracy versus communication/computation overheads ($K=80$). \label{subfig:various_EDs_d}]{\includegraphics[width=0.49\linewidth]{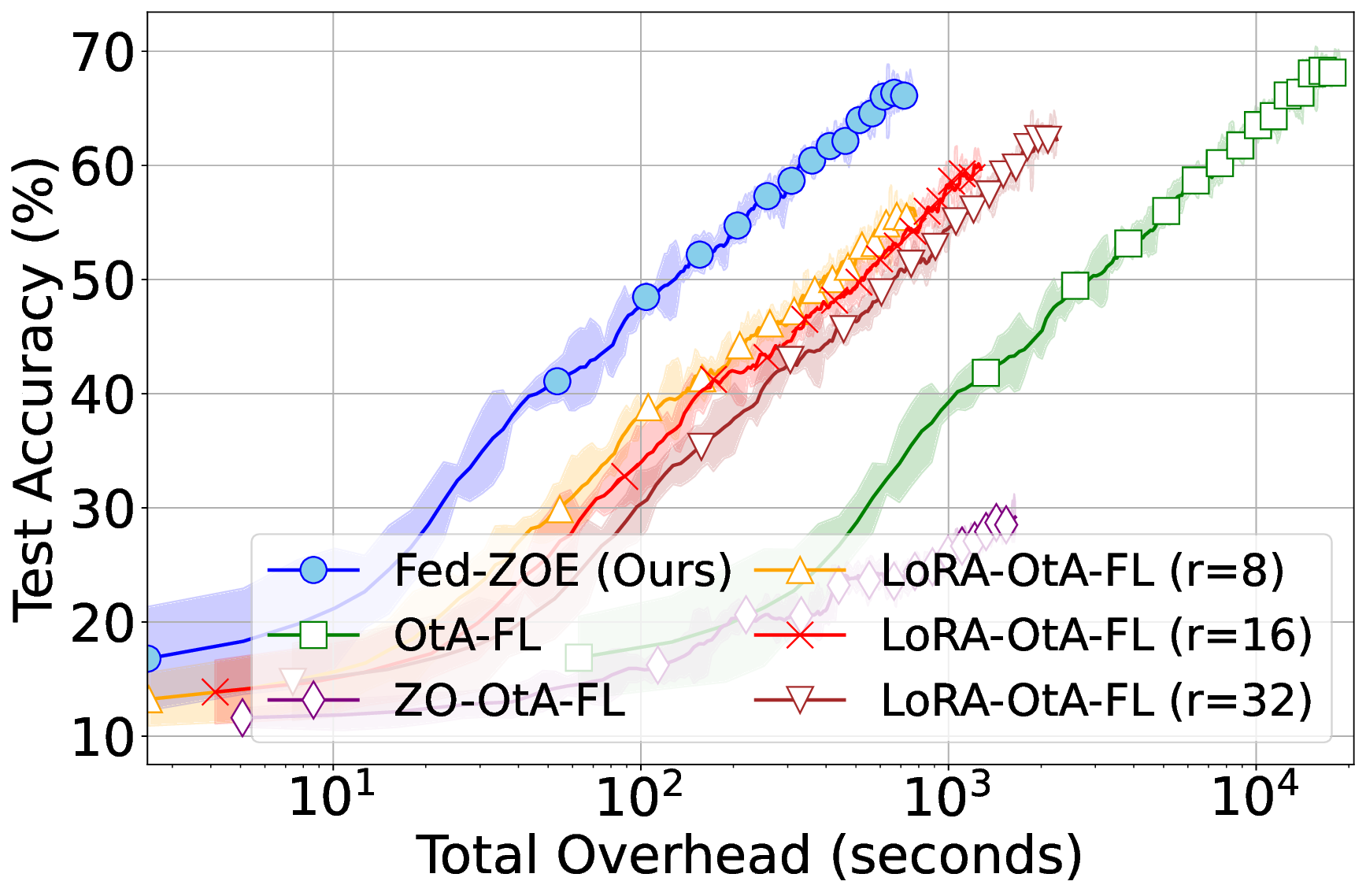}}\\
    \subfloat[Test accuracy versus communication rounds ($K=160$).\label{subfig:various_EDs_e}]{\includegraphics[width=0.49\linewidth]{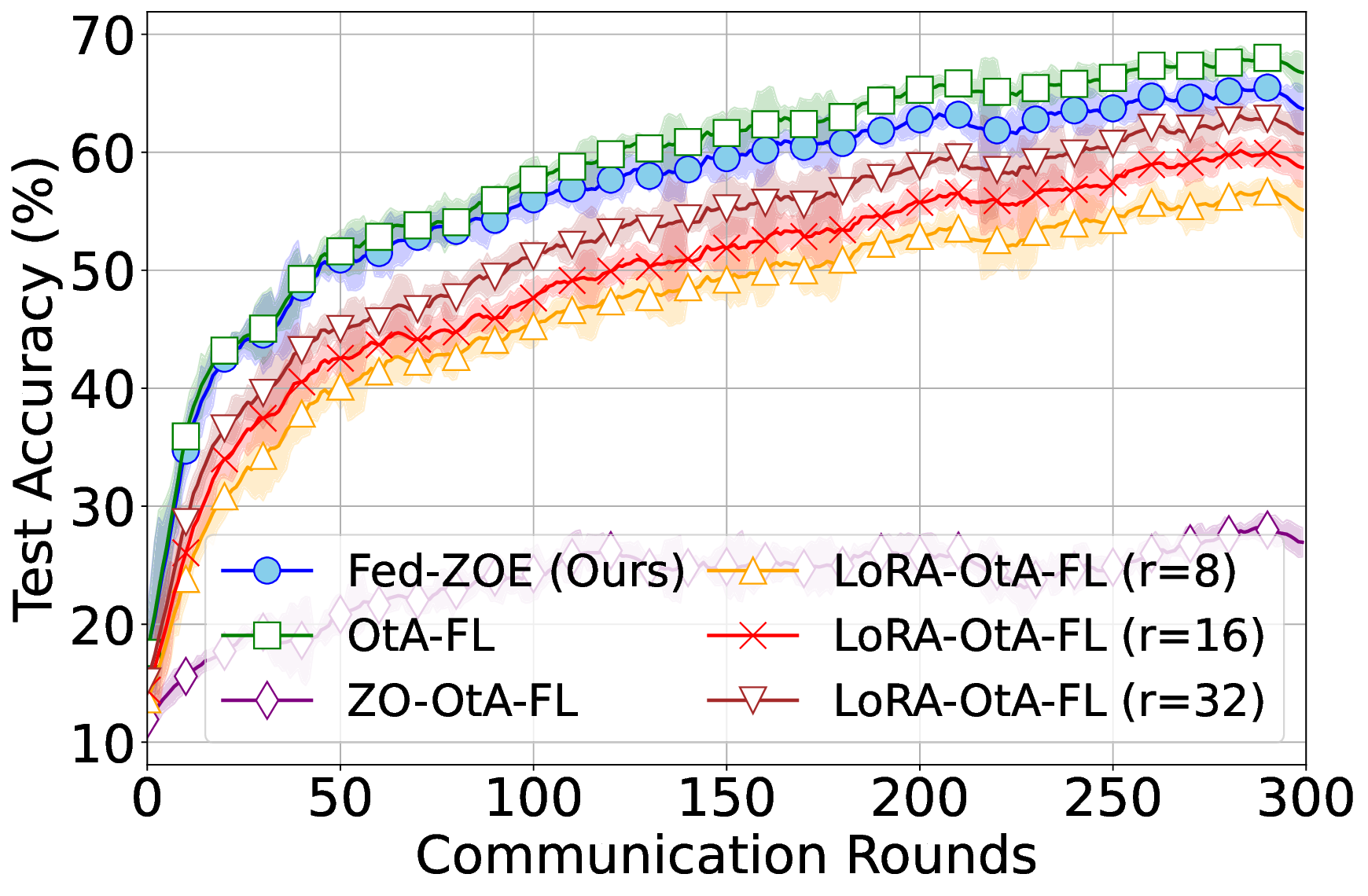}}
    \hfill
    \subfloat[Test accuracy versus communication/computation overheads ($K=160$). \label{subfig:various_EDs_f}]{\includegraphics[width=0.49\linewidth]{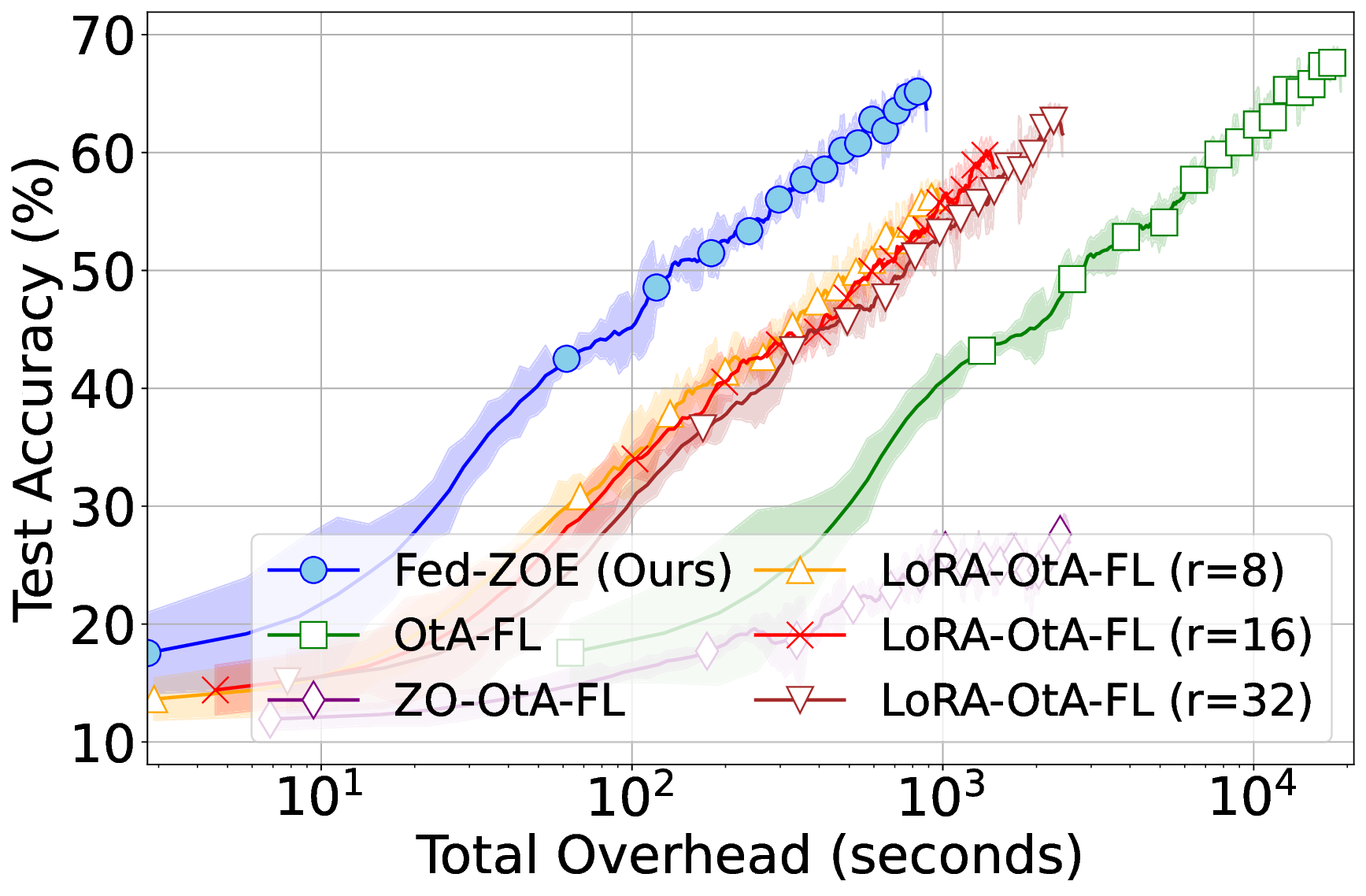}}
    \caption{Convergence evaluation of the proposed method and baseline methods with various numbers of EDs, $K\in\{20,80,160\}$. Here, we train a ResNet-18 model with the CIFAR-10 dataset.  \protect\subref{subfig:various_EDs_a}, \protect\subref{subfig:various_EDs_c},  \protect\subref{subfig:various_EDs_e}: the test accuracy is evaluated for each communication round. 
    \protect\subref{subfig:various_EDs_b}, \protect\subref{subfig:various_EDs_d}, \protect\subref{subfig:various_EDs_f}: the test accuracy is depicted versus the total loads.
    }
    \label{fig:various_EDs}
\end{figure}

\begin{figure*}
    \subfloat[Brain-CT dataset.\label{subfig:convergence_datasets_a}]{\includegraphics[width = 0.25\linewidth]{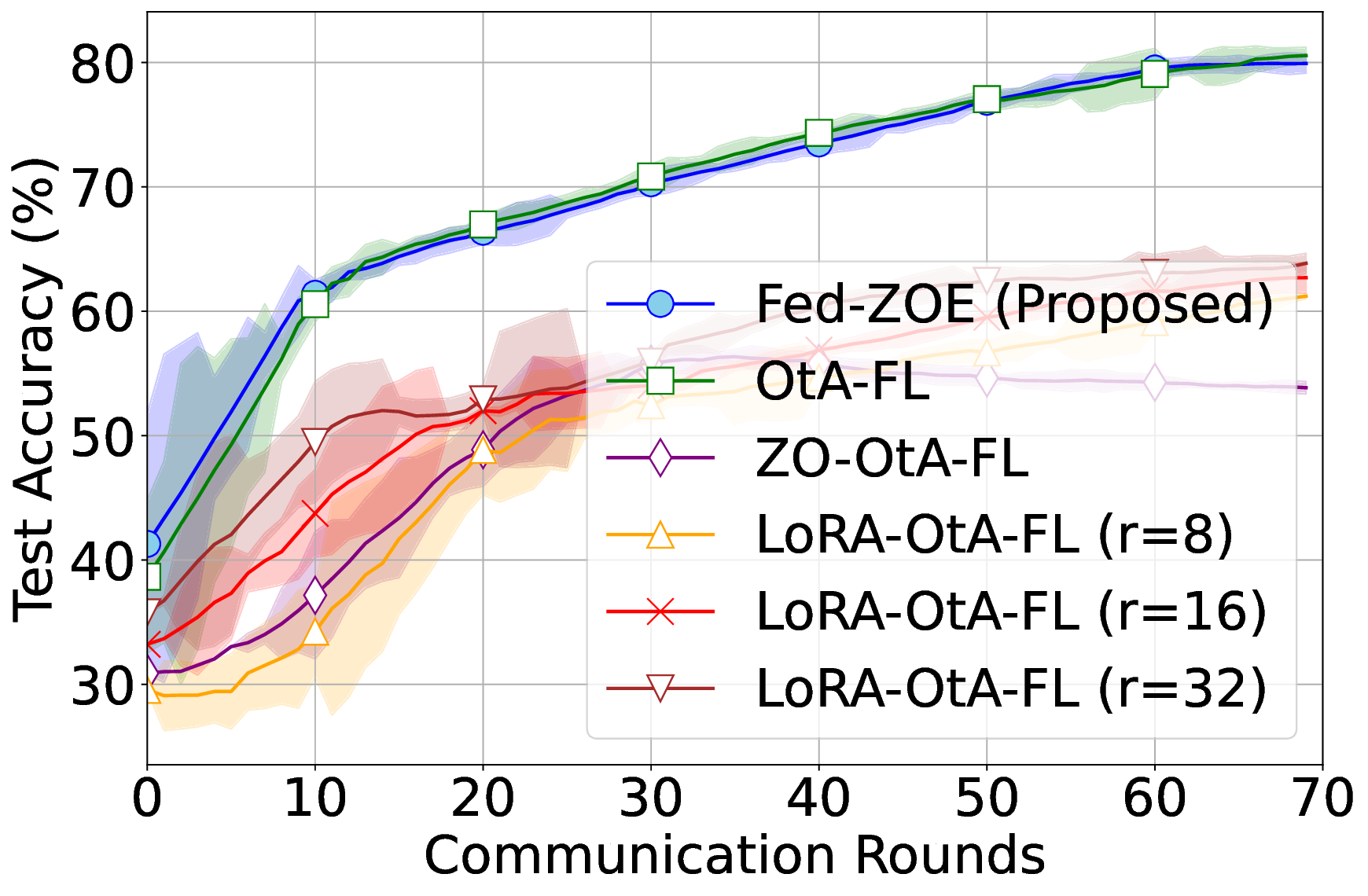}}
    \subfloat[SVHN dataset.\label{subfig:convergence_datasets_b}]{\includegraphics[width = 0.25\linewidth]{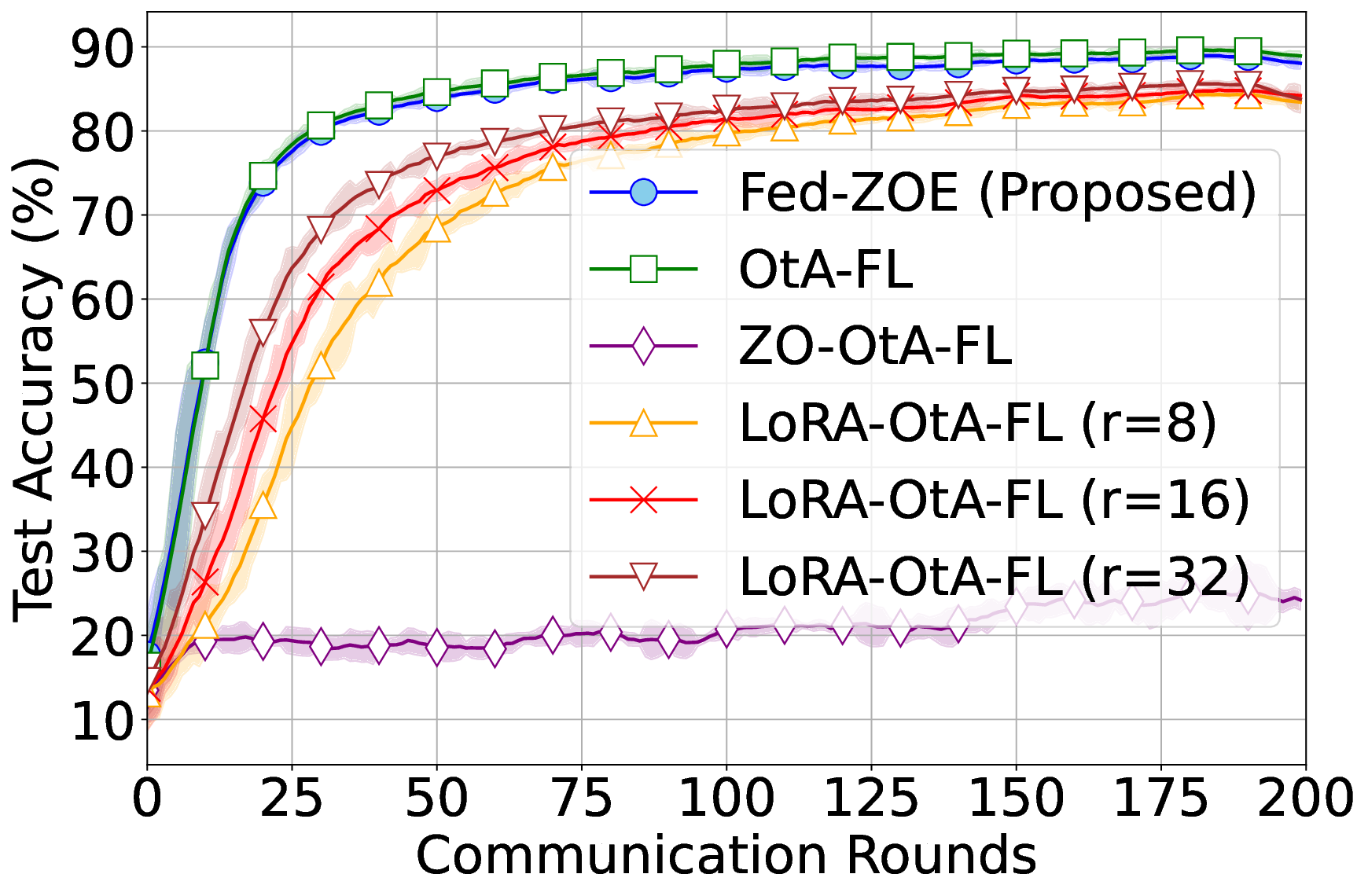}}
    \subfloat[Tiny-ImageNet dataset.\label{subfig:convergence_datasets_c}]{\includegraphics[width = 0.25\linewidth]{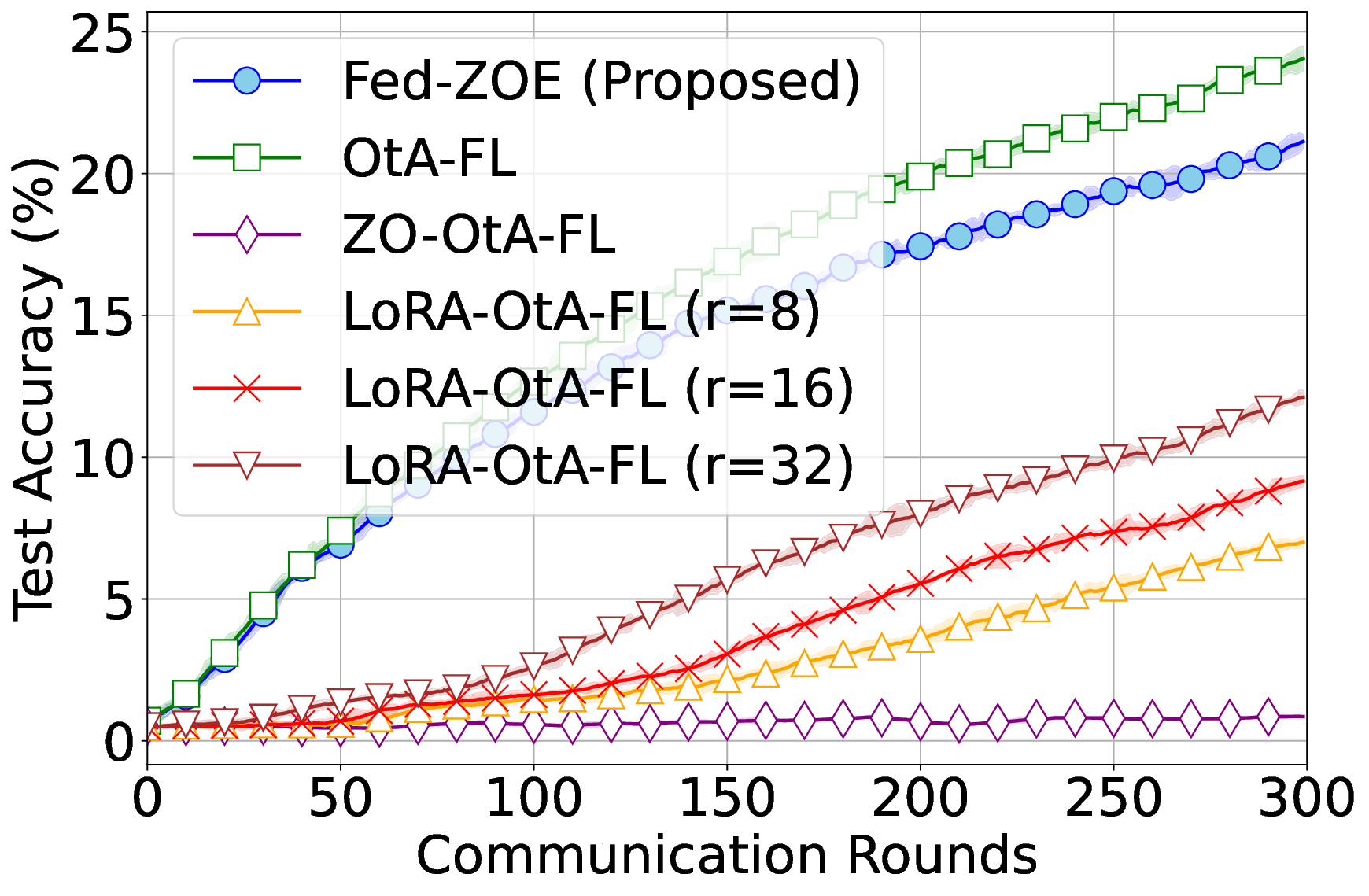}}
    \subfloat[CIFAR-100 dataset.\label{subfig:convergence_datasets_d}]{\includegraphics[width = 0.25\linewidth]{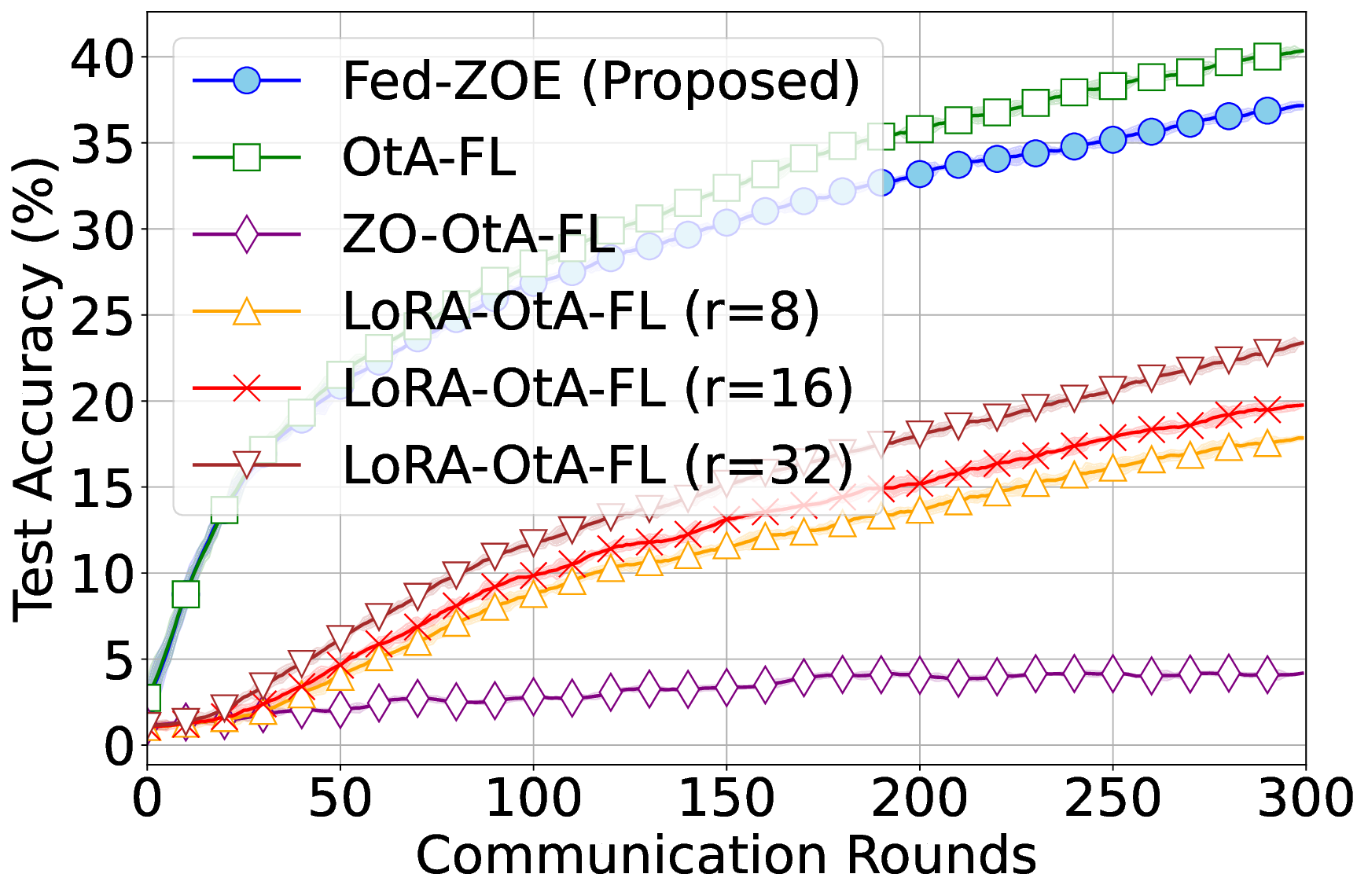}}
    \caption{Convergence of the FL methods for various datasets: \protect\subref{subfig:convergence_datasets_a} Brain-CT dataset, \protect\subref{subfig:convergence_datasets_b} SVHN dataset, \protect\subref{subfig:convergence_datasets_c} Tiny-ImageNet dataset, and \protect\subref{subfig:convergence_datasets_d} CIFAR-100 dataset.   The test accuracy is evaluated for every communication round. }\label{fig:convergence_datasets}
\end{figure*}

\begin{figure*}
    \subfloat[Brain-CT dataset.\label{subfig:load_datasets_a}]{\includegraphics[width = 0.25\linewidth]{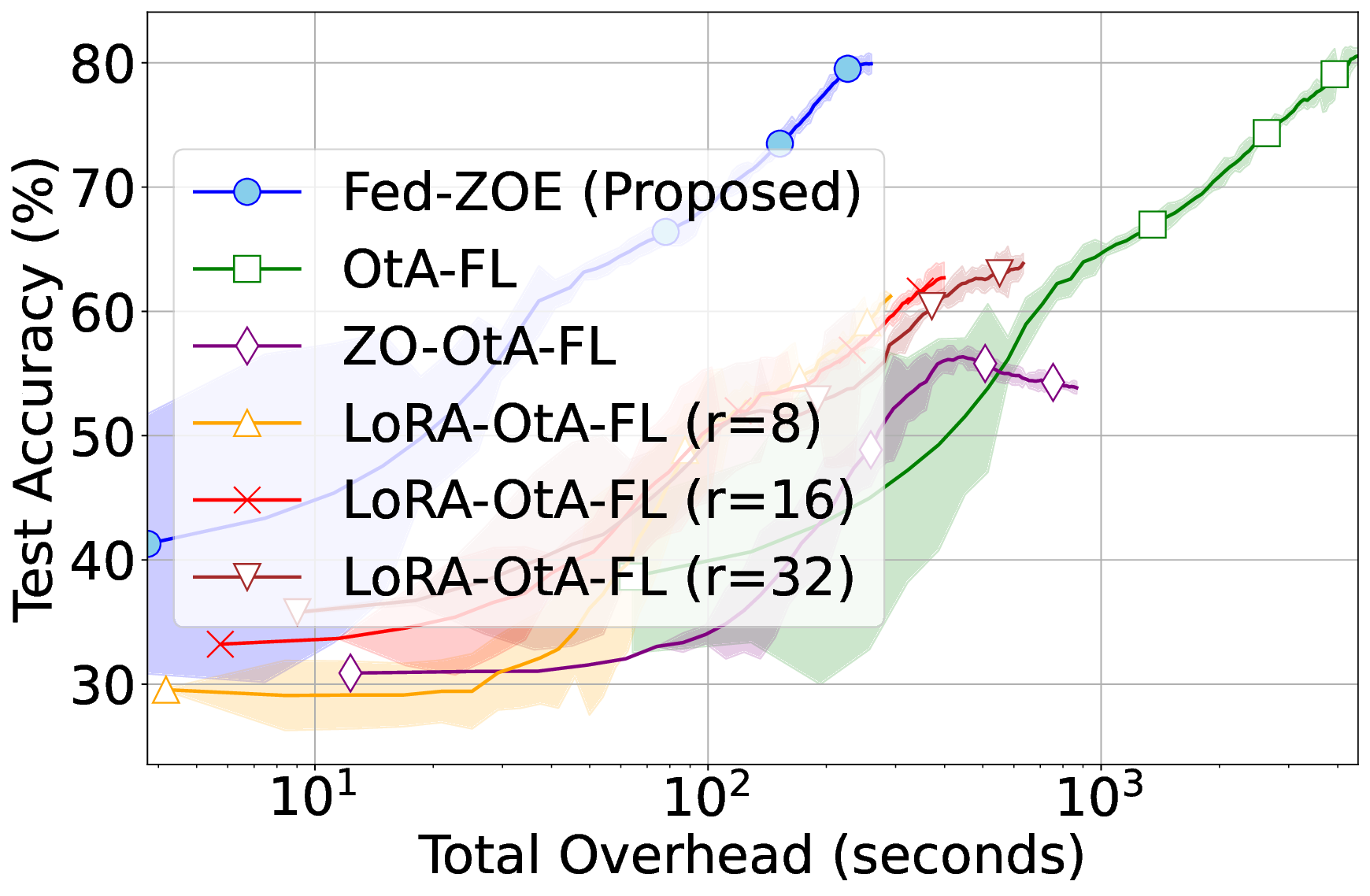}}
    \subfloat[SVHN dataset.\label{subfig:load_datasets_b} ]{\includegraphics[width = 0.25\linewidth]{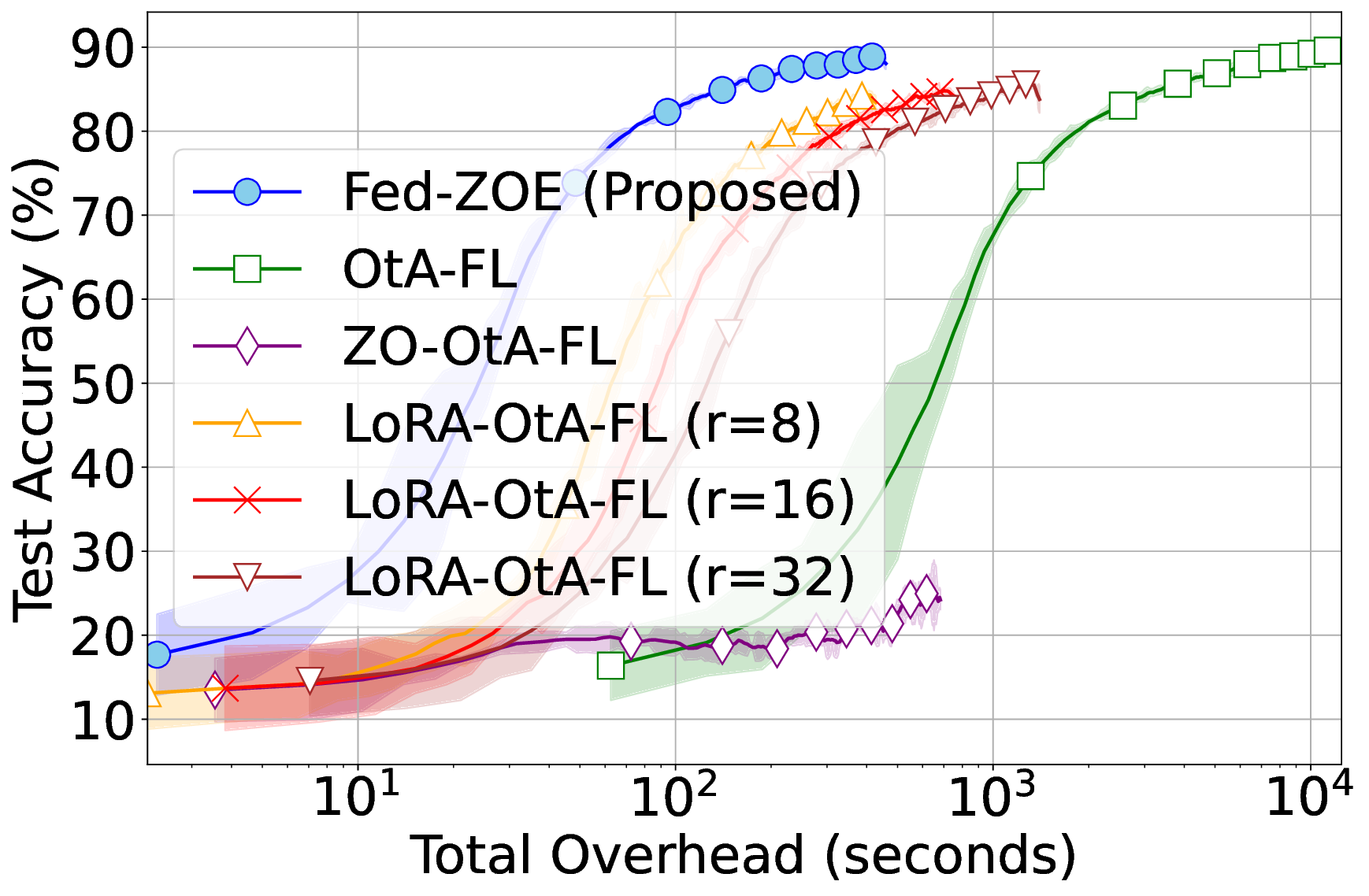}}
    \subfloat[Tiny-ImageNet dataset.\label{subfig:load_datasets_c}]{\includegraphics[width = 0.25\linewidth]{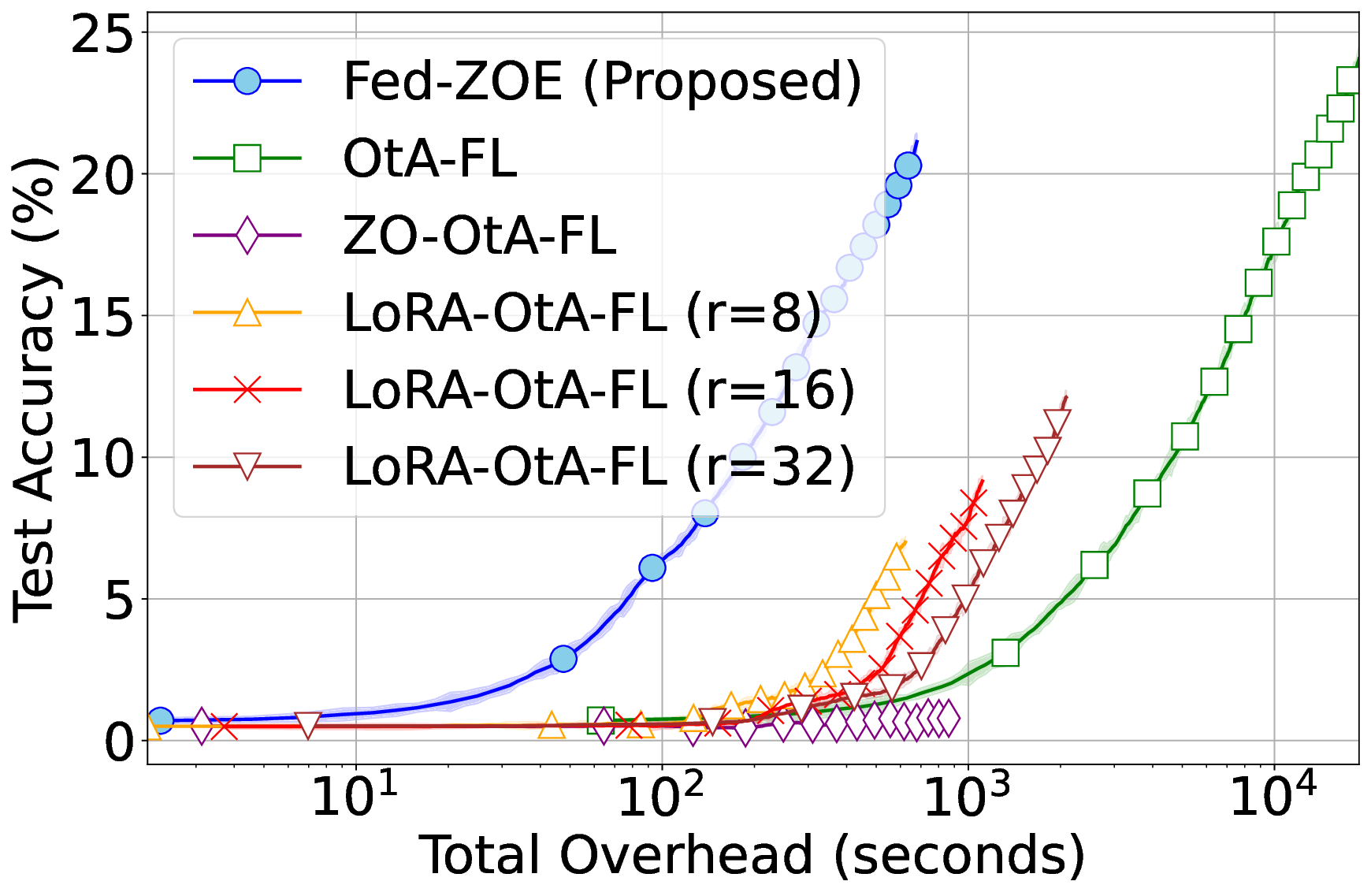}}
    \subfloat[CIFAR-100 dataset.\label{subfig:load_datasets_d} ]{\includegraphics[width = 0.25\linewidth]{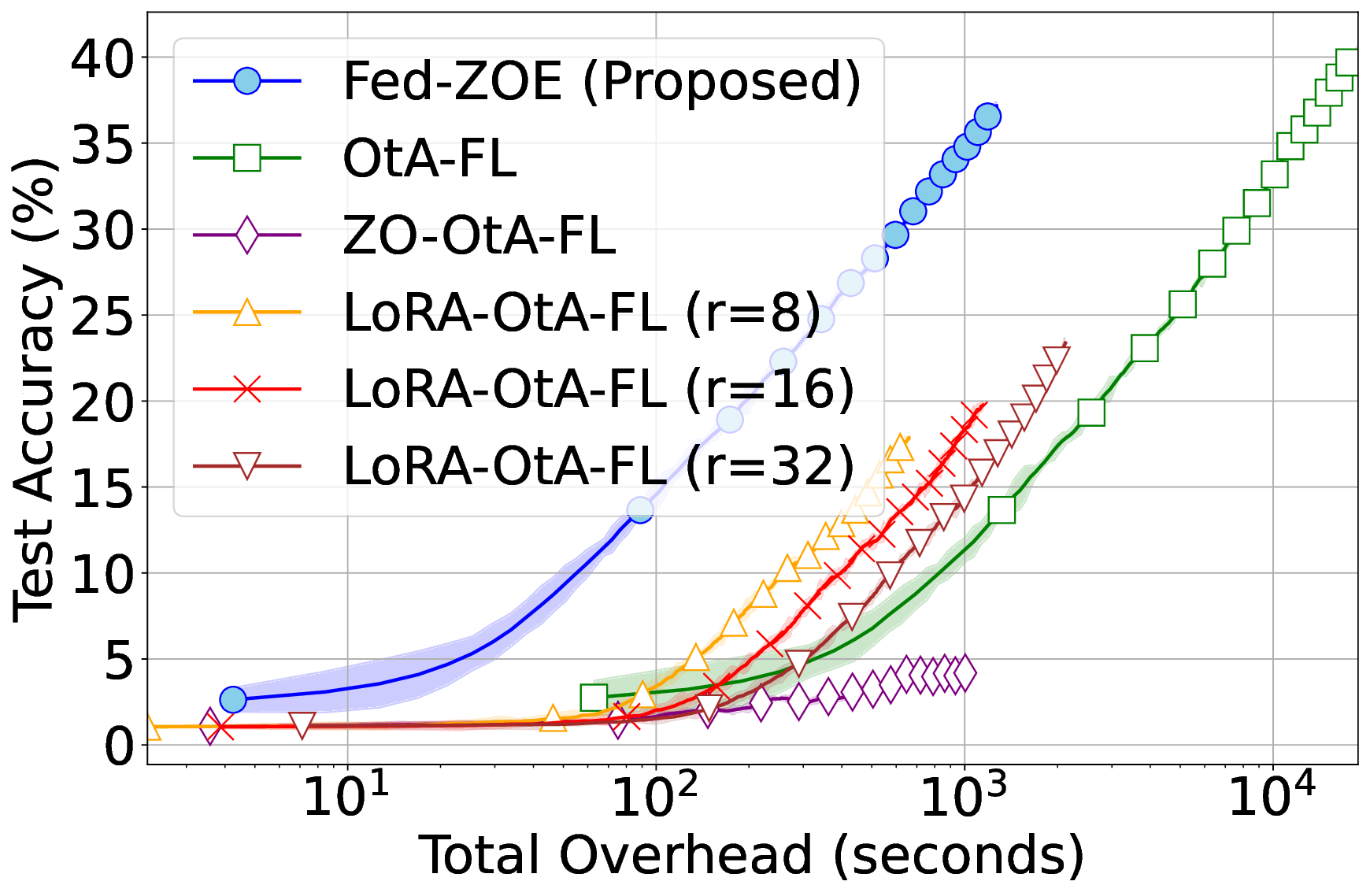}}
    \caption{Training efficiency of the FL methods for various datasets: \protect\subref{subfig:load_datasets_a} Brain-CT dataset, \protect\subref{subfig:load_datasets_b} SVHN dataset, \protect\subref{subfig:load_datasets_c} Tiny-ImageNet dataset, and \protect\subref{subfig:load_datasets_d} CIFAR-100 dataset. The test accuracy is depicted versus the total communication and computation loads. }\label{fig:load_datasets}
\end{figure*}

\paragraph*{Communication and Computation Loads}

Here, we provide an in-depth analysis of the computation and communication loads of the FL methods. 
Figure \ref{fig:loads_comparison} depicts the computation and communication per each communication round. 
Most importantly, the proposed method significantly reduces the communication overhead compared to the OtA-FL scheme, because the proposed method the gradient vectors into several scalar values. 
However, the compression procedure tasks an additional communication overhead of 8.8 (=1.79-0.9), which is negligible compared to the communication overhead of the OtA-FL scheme.
Thus, as shown in \cref{fig:convergence_cifar10_b}, the proposed method significantly outperforms the OtA-FL in efficiency. 
Also, LoRA-OtA-FL reduces the communication overhead by adding low-rank adaptors with slightly increased computation overhead; however, the communication overhead is heavier than that of the proposed method. 
Compared to the ZO-OtA-FL, the proposed method has a lighter computation overhead even though the gradient estimation of the proposed method is more accurate than the ZO-OtA-FL. 
By doing so, the proposed method significantly outperforms all the baselines in training efficiency.

\paragraph*{Various values of RGE random samples $L$}

Here, we compare the effect of the number of RGE samples $L$ on the convergence of the proposed method. 
To this end, we depict the convergence curve with various values of $L$ in \cref{fig:various_L_convergence}. 
In the figure, the test accuracy is gradually enhanced as the proposed method leverages more RGE samples. 
As shown in \cref{thm:convergence}, the more the RGE samples, the higher the convergence speed is. 

In addition to the convergence curve, we show the tradeoff between the communication/computation overhead and the test accuracy by varying the number of RGE samples $L$ in \cref{fig:various_samples}.
In the figure, we first scatter several points of (total loads, test accuracy) for $L\in\{512, 1024, 2048, 4096, 8192\}$. Second, we scatter the points of the baselines. 
As depicted, as $L$ increases, the test accuracy of the proposed method is enhanced by compromising the total load of the proposed method. 
With this figure, we confirm that the proposed method significantly reduces the total loads compared to the baseline methods.

\paragraph*{Various numbers of EDs $K$}

In \cref{fig:various_EDs}, we study the effect of the number of EDs.  
In \cref{subfig:various_EDs_a,subfig:various_EDs_c,subfig:various_EDs_e}, we depict the test accuracy for each communication round. 
By comparing these figures, we confirm that the test accuracy is gradually degraded as $K$ increases because the data distribution of EDs gets more heterogeneous.
Similarly, in \cref{subfig:various_EDs_b,subfig:various_EDs_d,subfig:various_EDs_f}, the training efficiency is also degraded due to the same reason in \cref{subfig:various_EDs_a,subfig:various_EDs_c,subfig:various_EDs_e}. 
Compared with the baselines, the proposed scheme has a considerable performance gap between LoRA-OtA-FL and ZO-OtA-FL, while closely achieving the test accuracy of OtA-FL, similar to previous results.



\subsection{Datasets Other Than CIFAR-10}

In \cref{fig:convergence_datasets}, we depict the convergence of the proposed method and baseline methods for various datasets. 
In this experiment, we adopt four different datasets other than the CIFAR-10 dataset: a) Brain-CT dataset, b) SVHN dataset, c) Tiny-ImageNet dataset, and d) CIFAR-100 dataset. 
For relatively easier tasks in \cref{subfig:convergence_datasets_a,subfig:convergence_datasets_b}, the proposed scheme closely achieves the test accuracy of the OtA-FL with significantly reduced overhead. 
The gap between the proposed method and the baseline methods widens in \cref{subfig:convergence_datasets_a}, because the agents have highly heterogeneous gradients due to the dataset's small size hindering the accurate gradient accumulation in low-rank approximation. 
In \cref{subfig:convergence_datasets_c,subfig:convergence_datasets_d}, as training a classifier for the given datasets is getting harder, the gap between the proposed method and baseline methods widens, thereby representing the fast convergence of the proposed method. 
On the other hand, we depict the training efficiency in \cref{fig:load_datasets}. 
In all the subfigures, the proposed method is the most efficient among the other FL methods.

\section{Conclusion}
\label{sec:conclusion}


In this study, we explored the efficiency of the OtA-FL. 
Specifically, we addressed the research question: ``\textit{How can we reduce communication overhead without resorting to time-consuming quantization or compressed sensing techniques, while still preserving the accuracy of the trained models?}.''
Our proposed method demonstrates an extremely improved training efficiency with test accuracy similar to that of the uncompressed OtA-FL, as shown by the experiments with various datasets. 
As well as the improved performance, we theoretically guarantee the convergence of the proposed method.
This work contributes to improving the efficiency of the OtA-FL without substantial accuracy degradation.
We believe our findings with guide future efforts in OtA-FL.


While this paper provides promising results in efficient OtA-FL, it is not without limitations. 
The primary limitation lies in the selection of the random direction in the RGE procedure. 
Although we use the randomized direction sampled from a random seed, there is room for enhancing the performance by 1) reusing the previous round's direction or 2) error feedback of the RGE process.

\appendices

\section{Beamforming Optimization \label{sec:appendix_beamforming}}

Problem \ref{eq:P2} is a well-known form of the single-group multi-cast beamforming problem~\cite{sidiropoulos2006transmit}, which is a kind of quadratically constrained quadratic programming (QCQP). 
This problem is well-known as a non-convex NP-hard problem, where there have been numerous solutions by semidefinite relaxation (SDR)~\cite{sidiropoulos2006transmit, li2011transmit}, successive linearization approximation (SLA)~\cite{tran2013conic}, and branch and bound (BB)~\cite{lu2017efficient}.

As presented in~\cite{lu2017efficient}, the SDR approach has low accuracy in the obtained solution, and the BB-based approach achieves global optimal solution consuming exponential computational complexity.
The SLA-based methods, however, can obtain the near-optimal solution with affordable computational complexity~\cite{yang2020federated}; hence, we use SLA for the optimization of Problem \ref{eq:P2}, a special case of convex-concave programming (CCP). 

\paragraph*{Solution of Problem \ref{eq:P2}}
For brevity, we express $\mathbf{r}$ in a real vector from:
\begin{equation}
        \mathbf{v} = 
        \begin{bmatrix}
        \mathtt{Re}\{\mathbf{r}\}\\
        \mathtt{Im}\{\mathbf{r}\}
        \end{bmatrix}
        \in\mathbb{R}^{2\cdot N}.
\end{equation}
Then, the constraint \eqref{eq:P2_constraint} is rewritten as follows:
\begin{equation}
    \begin{split}
        |\mathbf{r}^\mathrm{H}\mathbf{h}_k|^2  & = \left(\mathtt{Re}\{\mathbf{r}\}^\mathrm{T}\mathtt{Re}\{\mathbf{h}_k\} + \mathtt{Im}\{\mathbf{r}\}^\mathrm{T}\mathtt{Im}\{\mathbf{h}_k\}  \right)^2 \\
        & + \left(\mathtt{Im}\{\mathbf{r}\}^\mathrm{T}\mathtt{Re}\{\mathbf{h}_k\} - \mathtt{Re}\{\mathbf{r}\}^\mathrm{T}\mathtt{Im}\{\mathbf{h}_k\}  \right)^2 \\
        & = \mathbf{v}^\mathrm{T}\mathbf{Q}_k\mathbf{v} \le \frac{\sigma_k^2}{K^2P} = \gamma_k^2,
    \end{split}
\end{equation}
where
\begin{equation}
    \mathbf{Q}_k = 
    \begin{bmatrix}
        \mathtt{Re}\{\mathbf{h}_k\}\\
        \mathtt{Im}\{\mathbf{h}_k\}
    \end{bmatrix}
    \begin{bmatrix}
        \mathtt{Re}\{\mathbf{h}_k\}\\
        \mathtt{Im}\{\mathbf{h}_k\}
    \end{bmatrix}^\mathrm{T}
    +
    \begin{bmatrix}
        \mathtt{Im}\{\mathbf{h}_k\}\\
        -\mathtt{Re}\{\mathbf{h}_k\}
    \end{bmatrix}
    \begin{bmatrix}
        \mathtt{Im}\{\mathbf{h}_k\}\\
        -\mathtt{Re}\{\mathbf{h}_k\}
    \end{bmatrix}^\mathrm{T}.
\end{equation}
Then, without loss of generality, Problem \ref{eq:P2} can be rewritten as follows:
\begin{problem} \label{eq:P3}
    \begin{alignat}{3}
        \text{\ref{eq:P3}: } & & \min_{\mathbf{v}\in\mathbb{R}^{2N}} ~~ & \Vert\mathbf{\mathbf{v}}\Vert_2^2 \label{P3:obj} & \\
                         {}  & & ~\text{s.t. } ~~ & \mathbf{v}^\mathrm{T} \mathbf{Q}_k \mathbf{v} \ge \gamma_k^2,~ \forall k\in[K]. &\label{eq:P3_constraint}
    \end{alignat}
\end{problem}
Problem \ref{eq:P3} is a form of difference of convex (DC) programming as the objective function is a convex function, and the space of $\mathbf{v}$ satisfying the constraint \eqref{eq:P3_constraint} is a concave set. 
The CCP algorithm is a representative and iterative algorithm for solving DC programming by approximating the concave part with its first-order linear approximation. 
In CCP algorithm, we need to initialize a receive combiner vector $\mathbf{v}^{(0)}$, where the constraint \eqref{eq:P3_constraint} is satisfied. 
After that, the $e$-th iterative update is composed of the following steps:
\begin{enumerate}
    \item \textbf{Linearize the concave part: } 
    For each $k\in[K]$, the constraint \eqref{eq:P3_constraint} is linearized by 
    \begin{equation}\label{eq:linear_approx_constraint}
        {\mathbf{v}^{(e)}}^\mathrm{T} \mathbf{Q}_k \mathbf{v}^{(e)} -2{\mathbf{v}^{(e)}}^\mathrm{T} \mathbf{Q}_k \mathbf{v}+  \gamma_k^2 \le 0.
    \end{equation}
    Since $-\mathbf{v}^\mathrm{T}\mathbf{Q}_k\mathbf{v}$ is concave, its first-order approximation is a global overestimator, \ie, $\gamma_k^2 - \mathbf{v}^\mathrm{T}\mathbf{Q}_k\mathbf{v} \le {\mathbf{v}^{(e)}}^\mathrm{T} \mathbf{Q}_k \mathbf{v}^{(e)} - 2 {\mathbf{v}^{(e)}}^\mathrm{T} \mathbf{Q}_k \mathbf{v}+  \gamma_k^2$, for all $\mathbf{v}$. 
    \item \textbf{Solve the convex quadratic programming (QP): }
    \begin{problem} \label{eq:P4}
        \begin{alignat}{3}
            \text{\ref{eq:P4}: } & & \min_{\mathbf{v}\in\mathbb{R}^{2N}} ~~ & \Vert\mathbf{\mathbf{v}}\Vert_2^2 \label{P4:obj} & \text{~~~s.t.~} \eqref{eq:linear_approx_constraint}.
        \end{alignat}
    \end{problem}
    \item \textbf{Update $\mathbf{v}$ by: }
    Let us denote the solution of Problem \ref{eq:P4} at the $n$-th step as ${\mathbf{v}^{(e)}_\text{opt}}$, we have
    \begin{equation}
        \mathbf{v}^{(e+1)} = {\mathbf{v}^{(e)}_\text{opt}}.
    \end{equation}
    \item \textbf{Convergence criteria: } Check if $\Vert\mathbf{v}^{(e+1)} - \mathbf{v}^{(e)}\Vert_2^2$ is sufficiently small. 
\end{enumerate}

\section{Proof of Proposition \ref{prop:proposition_1}\label{sec:appendix_proposition1}}
\paragraph*{Expectation of the recovered gradient}

From the definition of $\Delta\widehat{\mathbf{w}}_{G,t}$ in \eqref{eq:def_delta_w}, we have 
\begin{align}
    & \mathbb{E}\left[\Delta\widehat{\mathbf{w}}_{G,t}\right] = \frac{1}{L} \mathbb{E}[\mathbf{U}_t \mathbf{y}_{\text{G}}] \\ 
    & = \frac{1}{KL} \sum_{k\in[K]}\sum_{l\in [L]} \underbrace{\mathbb{E}[\mathbf{u}_{t,l} \mathbf{u}_{t,l}^\mathrm{H}]}_{=\mathbf{I}_S} 
    \mathbb{E}[\Delta \mathbf{w}_{k,t}]  + \frac{1}{L}\sum_{l\in[L]} \underbrace{\mathbb{E}[\mathbf{r}^\mathrm{H}\mathbf{n} \mathbf{u}_{t,l}]}_{=\mathbf{0}} \nonumber \\ 
    & = \frac{1}{K}\sum_{k\in[K]} \mathbb{E}[\Delta\mathbf{w}_{k,t}] = - \frac{\eta}{K}\sum_{k\in[K]}\sum_{i\in[I]} \underbrace{\mathbb{E}[\mathbf{g}_{k,t,i-1}]}_{=\nabla F_k(\mathbf{w}_{k,t,i-1})}, \nonumber
\end{align}
where $\mathbf{I}_S$ denotes the identity matrix with dimension of $S\times S$.

\paragraph*{Variance of the recovered gradient}
Here, our focus is to derive the upper bound of the recovered gradient's variance. 
\begin{equation}\label{eq:appendix_mean_var}
    \begin{split}
        & \mathbb{E} \left[ \Vert\Delta \widehat{\mathbf{w}}_{G,t} - \Delta \mathbf{w}_{G,t}\Vert^2 \right] = \underbrace{\frac{1}{L^2} \mathbb{E} \left[ \Vert\mathbf{r}^\mathrm{H}\mathbf{n} \mathbf{u}_{t,l}\Vert^2 \right]}_{(A): \textbf{Part A}} \\ 
        & + \underbrace{\frac{1}{K^2L^2}\sum_{k\in[K]}\sum_{l\in[L]}\mathbb{E}\left[\left\Vert(\mathbf{u}_{t,l}\mathbf{u}_{t,l}^\mathrm{H}-\mathbf{I}_S) \Delta\mathbf{w}_{k,t} \right\Vert^2\right]}_{(B): \textbf{Part B}} 
    \end{split}
\end{equation}

\paragraph*{Part A}
The part A in \eqref{eq:appendix_mean_var} can be represented by 
\begin{equation}
    \begin{split}
        (A) & = \frac{1}{L^2} \mathbb{E}\left[(\mathbf{r}^\mathrm{H}\mathbf{n})^2 \mathbf{u}_{t,l}^\mathrm{H}\mathbf{u}_{t,l}\right]  = \frac{1}{L^2} S N_0\mathbb{E}[\Vert\mathbf{r}\Vert^2].
    \end{split}
\end{equation}
To ensure convergence, it is necessary for part A to decay with $\eta_t^2$. 
Hence, we define the solution $\mathbf{r}$ of Problem \ref{eq:P2} as $\mathbf{r}_\mathbbm{1}$, when $\sigma_{k,t} = 1$ for all $k\in[K]$. 
Then, if $\sigma_{k,t}\neq 1$, we have 
\begin{equation}
    \begin{split}
        \mathbb{E}[\Vert\mathbf{r}\Vert^2]& \le \Vert\mathbf{r}_\mathbbm{1}\Vert^2 \cdot \mathbb{E}[\max_k(\sigma_{k,t}^2)] \\ 
        & \overset{(a)}{\le} \Vert\mathbf{r}_\mathbbm{1}\Vert^2 \cdot\mathbb{E}[\max_{k}\Vert\Delta\mathbf{w}_{k,t}\Vert^2] \\ 
        & \le \Vert\mathbf{r}_\mathbbm{1}\Vert^2 \eta_t^2 \mathbb{E}\left[ \max_{k} \bigg\Vert\sum_{i\in [I]} \mathbf{g}_{t,k,i-1}\bigg\Vert^2  \right]\\ 
        & \le \Vert\mathbf{r}_\mathbbm{1}\Vert^2 \eta_t^2 I  \mathbb{E}\left[ \max_{k} \sum_{i\in [I]} \Vert \mathbf{g}_{t,k,i-1}\Vert^2\right]\\ 
        & \le \Vert\mathbf{r}_\mathbbm{1}\Vert^2 \eta_t^2 I^2 G_2^2,
    \end{split}
\end{equation}
where inequality (a) holds because  $\sigma_{k,t}^2\le \frac{1}{L}\Vert\boldsymbol{\phi}_{k,t} \Vert^2 = \frac{1}{L}\Vert \mathbf{U}_t\Delta\mathbf{w}_{k,t}\Vert^2  \le \Vert\Delta\mathbf{w}_{k,t}\Vert^2$.
Finally, we have
\begin{equation}
    \begin{split}
        (A) \le \frac{I^2 S N_0 \eta_t^2 \Vert\mathbf{r}_\mathbbm{1}\Vert^2 G_2^2}{L^2}.
    \end{split}
\end{equation}

\paragraph*{Part B}

The part B in \eqref{eq:appendix_mean_var} can be rewritten by 
\begin{equation}
\label{eq:parta_proposition1}
\begin{split}
    (B)& = \frac{1}{K^2L^2}\sum_{k\in[K]}\sum_{l\in[L]}\mathbb{E}\left[\left\Vert(\mathbf{u}_{t,l}\mathbf{u}_{t,l}^\mathrm{H}-\mathbf{I}_S) \Delta\mathbf{w}_{k,t} \right\Vert^2\right] \\ 
    & = \frac{1}{K^2L^2} \sum_{k\in[K]}\sum_{l\in[L]} \mathbb{E}\left[\Delta\mathbf{w}_{k,t}^\mathrm{H}(\mathbf{u}_{t,l}\mathbf{u}_{t,l}^\mathrm{H})^2\Delta\mathbf{w}_{k,t}\right]\\ 
    & ~~~ - \frac{1}{K^2L} \sum_{k\in[K]}\mathbb{E}[\Vert\Delta\mathbf{w}_{k,t}\Vert^2].
\end{split}
\end{equation}
By Isserlis' theorem~\cite{isserlis1918formula}, the $(i,j)$-th element of $(\mathbf{u}_{t,l}\mathbf{u}_{t,l}^\mathrm{H})^2$ can be derived by 
\begin{equation}
\label{eq:Isserlis}
    \begin{split}
        & [(\mathbf{u}_{t,l}\mathbf{u}_{t,l}^\mathrm{H})^2]_{ij} = \sum_{s\in[S]}\mathbb{E}\left[[\mathbf{u}_{t,l}]_i[\mathbf{u}_{t,l}]_s[\mathbf{u}_{t,l}]_{s}[\mathbf{u}_{t,l}]_j\right]\\ 
        & = \sum_{s\in[S]}\big(2\mathbb{E}\left[[\mathbf{u}_{t,l}]_{i}[\mathbf{u}_{t,l}]_{s}\right]\mathbb{E}\left[[\mathbf{u}_{t,l}]_{s}[\mathbf{u}_{t,l}]_{j}\right] \\
        &\hspace{30pt}+ \mathbb{E}\left[[\mathbf{u}_{t,l}]_{i}[\mathbf{u}_{t,l}]_{j}\right]\mathbb{E}\left[[\mathbf{u}_{t,l}]_{s}^2\right]\big)\\
        &= 3\delta_{ij}S,
    \end{split}
\end{equation}
where $\delta_{ij}$ denotes Kronecker delta function. 
Then, by substituting \eqref{eq:Isserlis} into \eqref{eq:parta_proposition1}, we have 
\begin{equation}
    \begin{split}
        (B) & = \frac{1}{K^2L}\sum_{k\in[K]} (3S-1)\mathbb{E}[\Vert\Delta\mathbf{w}_{k,t}\Vert^2].
    \end{split}
\end{equation}

\section{Proof of Theorem \ref{thm:convergence}\label{appendix:proof_convergence}}

Here, we prove the convergence of the proposed method.
For the brevity of the notation, we define the expression for average local update of EDs as 
\begin{equation}\label{eq:appendix_local_average}
    \bar{\mathbf{w}}_{G,t,i} = \frac{1}{K} \sum_{k\in[K]}\mathbf{w}_{k,t,i},
\end{equation}
where $\bar{\mathbf{w}}_{G,t,0} = \mathbf{w}_{G,t}$.
We note that the averaged local weight in \eqref{eq:appendix_local_average} is defined only for the convergence analysis, which is not shown in the practical implementation of Fed-ZOE. 

At the $t$-th communication round, the PS aggregates the local updates of $K$ EDs by 
\begin{equation}
    \mathbf{w}_{G,t+1} = \mathbf{w}_{G,t} + \Delta \widehat{\mathbf{w}}_{G,t},
\end{equation}
where $\Delta \widehat{\mathbf{w}}_{G,t}$ is defined in \eqref{eq:def_delta_w}.
Then, we can split $\Delta\widehat{\mathbf{w}}_{G,t}$ by following two parts: 
\begin{equation}\label{eq:appendix_convergence_1}
    \begin{split}
        & \mathbb{E}\left[F(\mathbf{w}_{G,t+1})-F(\mathbf{w}_{G,t}) \right] \\
        &= \underbrace{\mathbb{E}\left[F(\mathbf{w}_{G,t+1}) - F(\bar{\mathbf{w}}_{G,t,I})\right]}_{\text{\textbf{Part C}: Additive noise caused by Fed-ZOE}} \\ 
        &~~~~~+ \sum_{i\in[I]} \underbrace{\mathbb{E}\left[F(\bar{\mathbf{w}}_{G,t,i}) - F(\bar{\mathbf{w}}_{G,t,i-1}) \right]}_{\text{\textbf{Part D:} Local updates}}.
    \end{split}
\end{equation}

\paragraph*{Part C}
From the assumption A2 in Assumption \ref{assumption:1}, we have
\begin{equation}
\begin{split}\label{eq:partA_smooth}
        &\mathbb{E}\left[F(\mathbf{w}_{G,t+1})-F(\bar{\mathbf{w}}_{G,t,I})\right]\\
        &~~\le \mathbb{E}\left[ \nabla F (\bar{\mathbf{w}}_{G,t,I})^\mathrm{T}\left(\Delta\widehat{\mathbf{w}}_{G,t}-\Delta\mathbf{w}_{G,t}\right)\right]\\
        &~~~~~+ \frac{\beta}{2}\mathbb{E}\left[ \left\Vert\Delta\widehat{\mathbf{w}}_{G,t}-\Delta\mathbf{w}_{G,t}\right\Vert^2\right].
\end{split}
\end{equation}
Using the results in \cref{prop:proposition_1}, we have 
\begin{align}\label{eq:appendix_convergence_A}
        & \mathbb{E}\left[F(\mathbf{w}_{G,t+1})-F(\bar{\mathbf{w}}_{G,t,I})\right] \\ 
        & \le \underbrace{\frac{\beta(3S-1)}{2K^2L}\sum_{k\in[K]} \mathbb{E}[\Vert\Delta\mathbf{w}_{k,t}\Vert^2]}_{\textbf{Part C1}} + \underbrace{\frac{\beta I^2 S N_0 \eta_t^2 \Vert\mathbf{r}_\mathbbm{1}\Vert^2 G_2^2}{2L^2}}_{\textbf{Part C2}}. \nonumber
\end{align}

\paragraph*{Part D}
For each $i\in[I]$, the part D can be represented by 
\begin{equation}
    \begin{split}
    &\mathbb{E}\left[F(\bar{\mathbf{w}}_{G,t,i}) - F(\bar{\mathbf{w}}_{G,t,i-1}) \right] \\
    & ~~~ \le \eta_t \mathbb{E} \left[\nabla F(\bar{\mathbf{w}}_{G,t,i-1})^\mathrm{T} \left(\bar{\mathbf{w}}_{G,t,i}-\bar{\mathbf{w}}_{G,t,i-1}\right)\right]\\
    & ~~~~~ + \frac{\eta_t^2\beta}{2}\mathbb{E}\left[\left\Vert \bar{\mathbf{w}}_{G,t,i}-\bar{\mathbf{w}}_{G,t,i-1} \right\Vert^2\right] \\ 
    & ~~~ = - \frac{\eta_t}{K} \mathbb{E}\left[\nabla F (\bar{\mathbf{w}}_{G,t,i-1})^\mathrm{T}\sum_{k\in[K]}\nabla F_k(\mathbf{w}_{t,k,i-1})\right] \\
    & ~~~~~ + \frac{\eta_t^2\beta}{2K^2}\mathbb{E}\left[\left\Vert \sum_{k\in[K]} \mathbf{g}_{t,k,i-1}\right\Vert^2\right].
    \end{split}
\end{equation}
Because $-\mathbf{a}^\mathrm{T}\mathbf{b} = - \frac{1}{2}\Vert\mathbf{a}\Vert^2 - \frac{1}{2}\Vert\mathbf{b}\Vert^2 + \frac{1}{2}\Vert\mathbf{a-b}\Vert^2$ holds, we have
\begin{align}\label{eq:appendix_convergence_B}
    & (\textbf{Part D}) = -\underbrace{\frac{\eta_t}{2}\mathbb{E}\left[\Vert \nabla F(\bar{\mathbf{w}}_{G,t,i-1}) \Vert^2\right]}_{\textbf{Part D1}} \\ 
    & -\underbrace{\frac{\eta_t}{2} \mathbb{E}\bigg[\bigg\Vert \frac{1}{K}\sum_{k\in[K]} \nabla F_k(\mathbf{w}_{k,t,i-1})\bigg\Vert^2\bigg]}_{\textbf{Part D2}}\nonumber\\
    & + \underbrace{\frac{\eta_t}{2} \mathbb{E} \bigg[ \bigg\Vert \nabla F(\bar{\mathbf{w}}_{G,t,i-1}) - \frac{1}{K}\sum_{k\in[K]} \nabla F_k(\mathbf{w}_{t,k,i-1}) \bigg\Vert^2\bigg]}_{\textbf{Part D3}}\nonumber\\
    & + \underbrace{\frac{\eta_t^2\beta}{2K^2}\mathbb{E}\bigg[\bigg\Vert \sum_{k\in[K]} \mathbf{g}_{t,k,i-1}\bigg\Vert^2\bigg]}_{\textbf{Part D4}}, \nonumber
\end{align}

\paragraph*{Parts C1, D2 and D4}

Because $\Vert\sum_{i\in[I]}\mathbf{a}_n\Vert^2 \le I\sum_{i\in[I]} \Vert \mathbf{a}_n \Vert^2$, the \textbf{Part C1} is bounded by
\begin{equation}
    (\textbf{Part C1}) \le \frac{\eta_t^2\beta(3S-1)I}{2K^2L}\sum_{k\in[K]}\sum_{i\in[I]} \Vert \nabla F_k(\mathbf{w}_{t,k,i-1})\Vert^2.
\end{equation}
Also, \textbf{Part D4} except the coefficient $\frac{\eta_t^2\beta}{2K^2}$ is bounded by  
\begin{align}
        &\mathbb{E}\left[\left\Vert \sum_{k\in[K]} \mathbf{g}_{t,k,i-1}\right\Vert^2\right] = \mathbb{E}\left[\left\Vert \sum_{k\in[K]} \nabla F_k(\mathbf{w}_{t,k,i-1})\right\Vert^2\right] \nonumber \\ 
        & ~~~~~ + \mathbb{E}\left[\left\Vert \sum_{k\in[K]} (\mathbf{g}_{t,k,i-1} - \nabla F_k(\mathbf{w}_{t,k,i-1}))\right\Vert^2\right] \\
        & \le \mathbb{E}\left[\left\Vert \sum_{k\in[K]} \nabla F_k(\mathbf{w}_{t,k,i-1})\right\Vert^2\right] \nonumber \\ 
        & ~~~~~ +\underbrace{\sum_{k\in[K]}\mathbb{E} \left[\Vert \mathbf{g}_{t,k,i-1} - \nabla F_k(\mathbf{w}_{t,k,i-1}) \Vert^2\right]}_{\le K\xi^2 }.\nonumber
\end{align}
Because we choose learning rate $\eta_t^2\frac{\beta}{2}\left(1+\frac{(3S-1)I}{K^2L}\right)-\eta_t \le 0$, \ie, $\eta_t \le \frac{2}{\beta\left(1+\frac{(3S-1)I}{K^2L}\right)} $ in \cref{thm:convergence}, we have 
\begin{equation}\label{eq:appendix_convergence_A1}
    \begin{split}
        &\left(\eta_t^2\frac{\beta}{2}\left(1+\frac{(3S-1)I}{K^2L}\right)-\eta_t\right) \\
        & ~~~~~ \times \mathbb{E}\left[\left\Vert\sum_{k\in[K]}\frac{1}{K}\nabla F_k(\mathbf{w}_{t,k,i-1})\right\Vert^2\right] \le 0.
    \end{split}
\end{equation}

Finally, we can bound the sum of \textbf{Part C1}, \textbf{Part D2}, and \textbf{Part D4} by 
\begin{equation}
    (\textbf{Part C1}) + (\textbf{Part D2})+ (\textbf{Part D4}) \le \frac{\eta_t^2\xi^2\beta}{2K}.
\end{equation}

\paragraph*{Part D3}

In \cref{thm:convergence}, we aim to show the convergence in terms of $\frac{1}{TI}\sum_{t\in[T]}^{T-1}\sum_{i\in[I]}\Vert \nabla F(\bar{\mathbf{w}}_{G,t-1,i-1})\Vert^2$; thus, the only remaining term is \textbf{Part C2} and \textbf{Part D3}, where \textbf{Part C2} is a constant. 
\textbf{Part D3} is bounded as follows:
\begin{equation}
    \begin{split}
        &\frac{\eta_t}{2} \left\Vert \nabla F(\bar{\mathbf{w}}_{G,t,i-1}) - \sum_{k\in[K]} \frac{1}{K}\nabla F_k(\mathbf{w}_{t,k,i-1}) \right\Vert^2\\
        =&  \frac{\eta_t}{2} \left\Vert \sum_{k\in[K]}\frac{1}{K}\left(\nabla F_k(\bar{\mathbf{w}}_{G,t,i-1}) -  \nabla F_k(\mathbf{w}_{t,k,i-1}) \right)\right\Vert^2\\
        \le & \frac{\eta_t}{2}\sum_{k\in[K]}\frac{1}{K^2} \Vert \nabla F_k(\bar{\mathbf{w}}_{G,t,i-1}) -  \nabla F_k(\mathbf{w}_{t,k,i-1})\Vert^2.\\ 
    \end{split}
\end{equation}
Because we have assumed that the local loss function $F_k$ is a $\beta$-smoothness function for all $k\in[K]$, we have
\begin{align}
        &  \Vert \nabla F_k(\bar{\mathbf{w}}_{G,t,i}) -  \nabla F_k(\mathbf{w}_{t,k,i})\Vert^2 \nonumber 
        \le \beta^2 \Vert \bar{\mathbf{w}}_{G,t,i} - \mathbf{w}_{t,k,i}\Vert^2\nonumber \\ 
        = & \eta_t^2\beta^2 \left\Vert \sum_{m=1}^{i} \left(\sum_{k'\in[K]} \hspace{-8pt} \frac{\nabla F_{k'}(\mathbf{w}_{k',t,m})}{K} - \nabla F_k(\mathbf{w}_{k,t,m}) \right)\right\Vert^2 \nonumber \\ 
        \overset{(a)}{\le} & \eta_t^2\beta^2i \sum_{m=1}^{i}\left\Vert \sum_{k'\in[K]} \frac{\nabla F_{k'}(\mathbf{w}_{k',t,m})}{K} - \nabla F_k(\mathbf{w}_{k,t,m}) \right\Vert^2 \nonumber\\
        \overset{(b)}{\le} & 2 \eta_t^2\beta^2i \sum_{m=1}^{i}\Bigg(\left\Vert \sum_{k'\in[K]\backslash \{k\}}\frac{\nabla F_{k'}(\mathbf{w}_{k',t,m})}{K}\right\Vert^2 \nonumber \\
        & ~~~~~~~~~~~~~~~~  +\left\Vert \left(1-\frac{1}{K}\right)\nabla F_k(\mathbf{w}_{k,t,m}) \right\Vert^2\Bigg)\nonumber\\
        \le & 2 \eta_t^2\beta^2i \sum_{m=1}^{i}\Bigg(\frac{K-1}{K^2}\sum_{k'\in[K]\backslash \{k\}} \left\Vert \nabla F_{k'}(\mathbf{w}_{k',t,m})\right\Vert^2 \nonumber \\ 
        & ~~~~~~~~~~~~~~~~ +\left\Vert \left(1-\frac{1}{K}\right)\nabla F_k(\mathbf{w}_{k,t,m}) \right\Vert^2\Bigg)\nonumber\\
        \le & 2 \eta_t^2 \beta^2 i  \sum_{m=1}^{i}\left(\frac{(K-1)^2}{K^2} + \frac{(K-1)^2}{K^2} \right)G_1^2 \nonumber \\ 
        \le & 4\eta_t^2 \beta^2 i^2 G_1^2, 
\end{align}
where the inequalities (a) and (b) hold because $\Vert\sum_{i=1}^{n}\mathbf{a}_i\Vert^2\le n\sum_{i=1}^{n}\Vert\mathbf{a}_i\Vert^2$.
Finally, we have
\begin{equation}
    \begin{split}
         (\textbf{Part D3}) \le \frac{2 \eta_t^3 \beta^2 (i-1)^2 G_1^2}{K}. \label{eq:appendix_convergence_B3}
    \end{split}
\end{equation}

\vspace{-10pt}

\paragraph*{Results}
From \cref{eq:appendix_convergence_1,eq:appendix_convergence_A,eq:appendix_convergence_B,eq:appendix_convergence_A1,eq:appendix_convergence_B3}, we have
\begin{align}
        & \mathbb{E}\left[F(\mathbf{w}_{G,t+1})-F(\mathbf{w}_{G,t}) \right] \nonumber\\
        & \le - \frac{\eta_t}{2}\sum_{i\in[I]}\mathbb{E}[\Vert \nabla F(\bar{\mathbf{w}}_{G,t,i-1})\Vert^2]  + \frac{\eta_t^2\xi^2\beta}{2K} \\ 
        & ~  +\frac{\beta I^2 S N_0 \eta_t^2 \Vert\mathbf{r}_\mathbbm{1}\Vert^2 G_2^2}{2L^2} + \frac{2 \sum_{i\in[I]}\eta_t^3 \beta^2 (i-1)^2 G_1^2}{K}\nonumber .  
\end{align}
Summing up from $t=0$ to $t=T-1$, we have
\begin{align}\label{eq:appendix_convergence_final}
    & \mathbb{E}\left[F(\mathbf{w}_{G,T}) - F(\mathbf{w}_{G,0})\right] \\
    & \le \frac{\beta T I^2 S N_0 \eta_t^2 \Vert\mathbf{r}_\mathbbm{1}\Vert^2 G^2}{2L^2}  - \sum_{t=0}^{T-1}\sum_{i\in[I]}\frac{\eta_t}{2}\Vert \nabla F(\bar{\mathbf{w}}_{G,t,i-1})\Vert^2 \nonumber\\ 
    & ~~~~~ + \frac{2 T \eta_t^3 \beta^2 (I-1)I^2 G^2}{3K} + \frac{T \eta_t^2\xi^2\beta}{2K}, \nonumber
\end{align}
where the inequality holds because $\sum_{i\in[I]} (i-1)^2 \le \frac{(I-1)I^2}{3}$.
By substituting $\eta_t=\frac{1}{\sqrt{TI}}\le \frac{2}{\beta\left(1+\frac{(3S-1)I}{K^2L}\right)} $ into \eqref{eq:appendix_convergence_final}, we have
\begin{equation}
    \begin{split}
        &\frac{1}{TI}\sum_{t\in[T]}\sum_{i\in[I]}  \mathbb{E}\left\Vert \nabla F(\bar{\mathbf{w}}_{G,t-1,i-1})\Vert^2\right] \\
        &\le \frac{2}{\sqrt{TI}}\mathbb{E}\left[F(\mathbf{w}_{G,0}) - F(\mathbf{w}_{G,T})\right] \\
        & + \frac{\beta \sqrt{I} S N_0  \Vert\mathbf{r}_\mathbbm{1}\Vert^2 G_2^2}{\sqrt{T}L^2} +  \frac{ 4 \beta^2 (I-1) G_1^2}{3TK} + \frac{\xi^2\beta}{2KI^\frac{3}{2}\sqrt{T}}. 
    \end{split}
\end{equation}

\bibliographystyle{IEEEtran}
\bibliography{main}

\end{document}